\pdfoutput=1

\documentclass[11pt]{article}

\usepackage[preprint]{acl}

\usepackage{times}
\usepackage{latexsym}

\usepackage[T1]{fontenc}

\usepackage[utf8]{inputenc}

\usepackage{microtype}

\usepackage{inconsolata}
\usepackage{hyperref}
\usepackage{graphicx}
\usepackage{booktabs}
\usepackage{multirow}
\usepackage{url}
\usepackage{pgfplotstable}
\usepackage{array}
\usepackage{caption}
\usepackage{subcaption}
\usepackage{authblk}
\usepackage{rotating}
\usepackage[originalparameters]{ragged2e}
\pgfplotsset{compat=1.18}

\title{Too Big to Fail: Larger Language Models are Disproportionately Resilient to Induction of Dementia-Related Linguistic Anomalies}

\author[1]{Changye Li}
\author[1]{Zhecheng Sheng}
\author[2]{Trevor Cohen}
\author[1]{Serguei Pakhomov}

\affil[1]{University of Minnesota}
\affil[2]{University of Washington}

\affil[1]{\texttt{\{lixx3013, sheng136, pakh0002}\}@umn.edu} 
\affil[2]{\texttt{\{cohenta}\}@uw.edu}

\begin{document}
\maketitle
\begin{abstract}
As artificial neural networks grow in complexity, understanding their inner workings becomes increasingly challenging, which is particularly important in healthcare applications. The intrinsic evaluation metrics of autoregressive neural language models (NLMs), perplexity (PPL), can reflect how ``surprised'' an NLM model is at novel input. PPL has been widely used to understand the behavior of NLMs.
Previous findings show that changes in PPL when masking attention layers in pre-trained transformer-based NLMs reflect linguistic anomalies associated with Alzheimer's disease dementia. Building upon this, we explore a novel bidirectional attention head ablation method that exhibits properties attributed to the concepts of cognitive and brain reserve in human brain studies, which postulate that people with more neurons in the brain and more efficient processing are more resilient to neurodegeneration. Our results show that larger GPT-2 models require a disproportionately larger share of attention heads to be masked/ablated to display degradation of similar magnitude to masking in smaller models. These results suggest that the attention mechanism in transformer models may present an analogue to the notions of cognitive and brain reserve and could potentially be used to model certain aspects of the progression of neurodegenerative disorders and aging.

\end{abstract}

\section{Introduction}

Alzheimer's disease (AD) dementia is a currently incurable neurodegenerative condition that leads to a progressive and irreversible decline in cognitive function. Due to the challenging nature of early diagnosis of this condition, there is a pressing need for efficient and cost-effective screening tools \citep{bradford2009missed} to mitigate the negative consequences of delayed or absent diagnosis \citep{doi.org/10.1111/psyg.12095}. Previous studies have demonstrated that changes in cognitive status can be reflected in spoken language and spontaneous speech \citep{doi:10.1080/02687039608248419, ALMOR1999202, HIER1985117}. Automated analysis of such speech, employing supervised machine learning models, shows its potential as an early screening tool. These models can be trained to identify subtle linguistic anomalies associated with dementia from transcripts of both healthy individuals and those with dementia. Recent advances in machine learning, such as deep learning models and the transformer with attention architecture \citep{NIPS2017_3f5ee243}, have mediated remarkable performance on this downstream task (for a review, see \citet{SHI2023115538}). Deep learning models, inspired by the human brain, are artificial neural networks (ANNs) that process vast amounts of data and learn complicated patterns, making them well-suited for analyzing subtle linguistic patterns. The transformer architecture, in particular, has advanced performance on natural language processing (NLP) tasks by enabling models to capture long-range dependencies more effectively via the attention mechanism \citep{NIPS2017_3f5ee243}.

As ANNs get larger and more complicated, it becomes even harder to interpret their inner workings. The performance of autoregressive neural language models (NLMs) (e.g., predicting the next word given the context) is frequently estimated with a single somewhat interpretable feature, perplexity (PPL), which has shown to be a suitable measurement for evaluating cognitive impairment from spontaneous speech \citep{fritsch2019automatic, cohen-pakhomov-2020-tale}. As the name ``perplexity'' suggests, it can be considered as an indicator of how ``surprised'' a model is by novel (i.e., not used in model's training) input. The more different the input is from a particular model's training data, the ``harder'' it is for the model to predict, resulting in higher PPL. Therefore, it is reasonable to hypothesize that PPL may have some degree of diagnostic utility, as an indicator of patterns of language use that fall outside the scope of the typical language used to train a model. In the context of AD, changes in language and cognitive function often manifest as differences in language complexity, with individuals experiencing difficulty in forming coherent sentences and selecting appropriate words. As AD progresses, the language used by patients with dementia becomes more unpredictable and less coherent, leading to higher PPL with models trained on language from individuals presumed to be cognitively healthy. 

While training data from cognitively healthy individuals is plentiful, language data produced by patients with dementia is much more impractical to obtain in sufficient quantity to train a large NLM. In hyperdimensional computing \citep{kanerva2009hyperdimensional}, high-dimensional vector representations are manipulated using operators that alter their distance from other learned representations. A prior work inspired by this concept \citep{li-etal-2022-gpt} demonstrates that masking the attention sub-modules of pre-trained transformer-based NLMs and thereby artificially increasing PPL on text from cognitively healthy individuals, can provide an effective solution to the challenge of limited data availability. By strategically altering these sub-modules and introducing controlled perturbations in the NLMs' attention layers, the degraded NLMs induce the linguistic anomalies and unpredictability associated with dementia.

Recent work in neuroscience using functional magnetic resonance imaging (fMRI) and electrocorticography (ECoG) has demonstrated that NLM's PPL is associated with predicting neural activation patterns during language comprehension tasks in the human brain \citep{doi:10.1073/pnas.2105646118, hosseini2024artificial}. This suggests a potential connection between the predictive capabilities of these models and understanding human information processing. In particular, one of the less well-understood phenomena in how neurodegeneration affects the human brain is the notion of cognitive and brain reserve. This notion is hypothesized to be responsible for findings that indicate individuals with higher innate abilities and/or aspects of life experience, such as educational and professional attainment, are able to mask the effects of dementia longer than those without these characteristics \citep{stern2002cognitive, stern2009cognitive, stern2012cognitive, scarmeas2004cognitive, scarmeas2003cognitive, snowdon1996linguistic}. In some cases, the notion of cognitive and brain reserve may even allow individuals to revert from initial signs of cognitive impairment to normal function \citep{iraniparast2022cognitive}.

Building upon these findings, our study seeks to further explore the potential of probing pre-trained GPT-2 family models \citep{radford2019language} to simulate cognitive impairment observed in patients with dementia with a specific focus on the cognitive reserve hypothesis. Using a set of transcripts from a widely-used ``Cookie Theft'' picture description cognitive task, we propose that the impaired information processing as the disease progresses can be simulated by masking a certain share of attention heads in a pre-trained GPT-2 model. Specifically, we follow the previously established paired-perplexity paradigm \citep{li-etal-2022-gpt} using a pair of unmasked (``control'') and masked (``dementia'') NLMs. In this approach, the difference between PPLs produced by these two NLMs is used to discriminate between picture descriptions by patients with dementia and healthy controls. We hypothesize that larger GPT-2 models with more attention heads will exhibit greater resilience to masking (i.e., a proxy for neural degeneration), necessitating a larger share of attention heads to be masked to achieve comparable classification performance to smaller models. We evaluate this hypothesis by targeting two subsets of attention heads that are a) \textit{most} important, and b) \textit{least} important to representation of the content of the ``Cookie Theft'' task, in which the degree of importance is ranked by the gradient changes in each attention head during fine-tuning of a pre-trained GPT-2 model to the content of the ``Cookie Theft'' transcripts. 

The contributions of this work can be summarized as follows: a) we provide preliminary evidence suggesting that the concept of cognitive reserve observed in human cognition appears to have an analog in ANNs; and b) our attention masking approach achieves comparable classification performance to another approach developed in prior work that directly artificially degrades NLM parameters \citep{li-etal-2022-gpt}, and the state-of-the-art (SOTA) model trained from scratch \citep{taghibeyglou2024context} with \textit{significantly fewer} trainable parameter masking/fitting.\footnote{The code to reproduce the results presented in this paper is available at \href{https://github.com/LinguisticAnomalies/artificial-neural-reserve}{GitHub}. The data are also publicly available but cannot be redistributed and must be obtained directly from Dementia Bank.}

\section{Background}

\subsection{Cognitive Reserve}

\begin{figure}[htbp]
\centering
\includegraphics[width=\columnwidth]{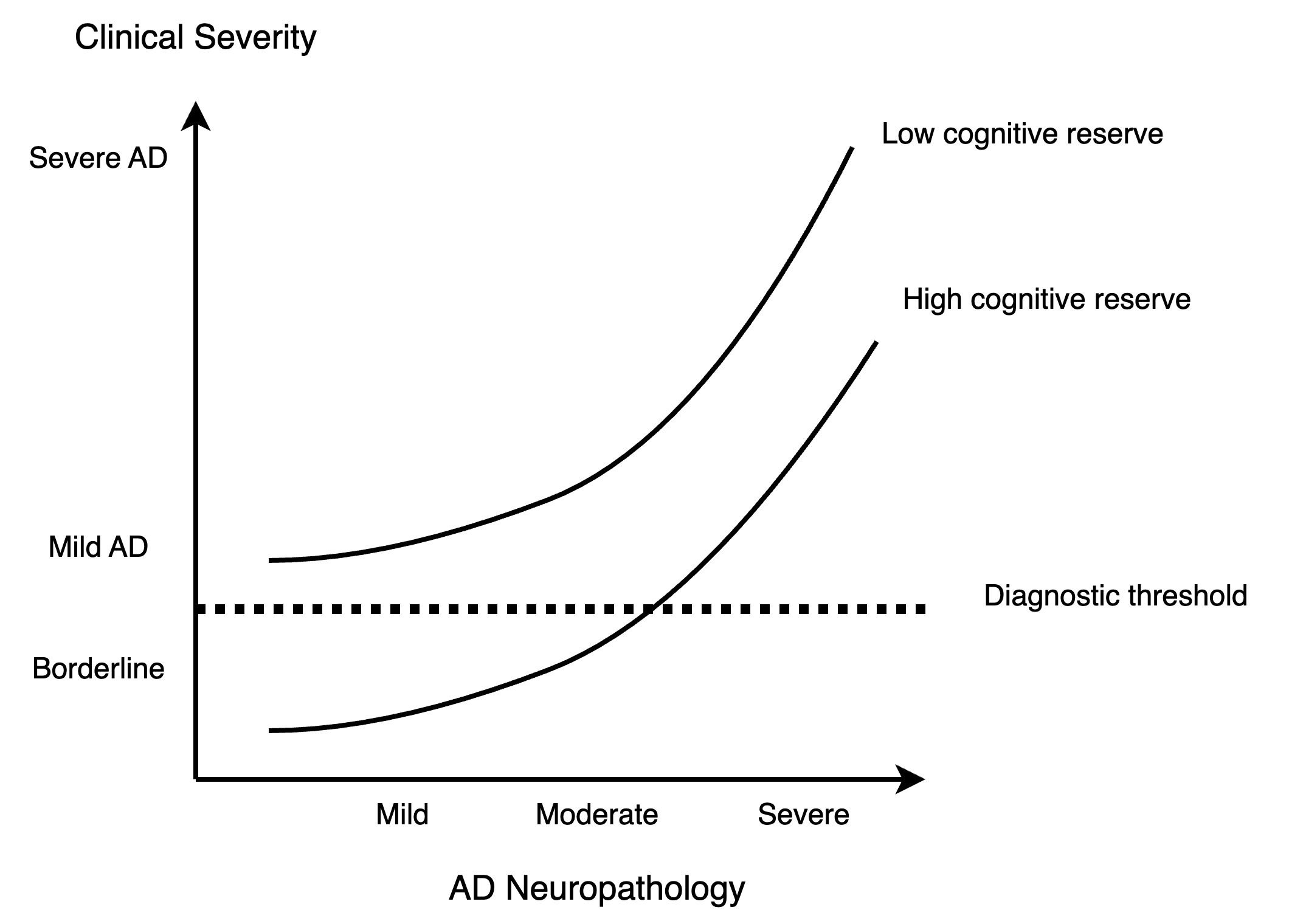}
\caption{A theoretical illustration of cognitive reserve and its mediation effect between AD neuropathology (x-axis) and clinical outcome (y-axis). Illustration derived from \citet{stern2002cognitive, stern2009cognitive}. As the disease progresses (i.e., with more impairment), individuals with higher cognitive/brain reserve would be more resilient to the effects, resulting in a lower level of clinical severity.}
\label{fig:cr}
\end{figure}

The notions of brain plasticity in the human brain and ``graceful degradation'' in ANNs have been extensively investigated in the neuroscientific literature demonstrating, for example, that a large proportion (over 80\%) of the connections in an ANN trained to simulate the motor cortex to generate signals directing body movement have to be ablated before the model's performance begins to collapse \citep{lukashin1994overlapping, lukashin1994neural}. The concepts of cognitive and brain reserves are closely related to brain plasticity applied to observations in neurodegenerative diseases as illustrated in Figure~\ref{fig:cr}. One of the earlier observations of this phenomenon comes from the Nun Study which found that low linguistic ability early in life (possibly due to innate abilities or educational attainment) is predictive of poor cognitive function and AD later in life \citep{snowdon1996linguistic}. The concept of cognitive reserve was further developed based on observations of the individual differences in effects of brain damage or pathology on clinical manifestations of cognitive function \citep{stern2002cognitive, stern2009cognitive, stern2012cognitive}. A multi-site study \citep{esiri2001pathological} reported that up to 25\% older adults without signs of cognitive impairment during neuropsychological testing meet all the histopathological criteria for AD (amyloid plaques and tau protein tangles) prior to their death. While this study did not assess brain volume, another similar study did find that a subgroup of 10 study participants who had both AD pathology and preserved mental status had greater brain weights and number of neurons in their brains \citep{katzman1988clinical}. 

A distinction can be made between the closely related notions of cognitive reserve and brain reserve. Cognitive reserve refers to the efficiency of brain networks, which manifests as greater educational and professional attainment. Brain reserve, on the other hand, refers to the physical properties of the brain, such as a larger number of neurons in biological neural network(s). This can manifest, for example, as a higher intelligence quotient. These two notions are difficult to disentangle due to their significant interdependence \citep{steffener2012exploring}. The properties of these notions have also been described using passive or active models that correspond to the notions of brain and cognitive reserves, respectively. Passive models \citep{katzman1993education, satz1993brain} measure the cognitive reserve by the size of the brain or the count of neurons in the brain. Passive models hypothesize that there is a threshold for brain reserve capacity - once an individual passes the ``point of no return'', the manifestation of neurodegenerative disease, such as AD, will occur regardless. Contrary to passive models, active models \citep{stern2002cognitive} hypothesize that there is a neural compensatory effect for brain damage. This effect consists of the brain compensating for the damage by activating other biological neural network(s) to perform cognitive task-related activities. In this case, patients of similar brain impairment but with more cognitive reserve may be more resilient to the disease's progression before the clinical manifestations of neurodegeneration become apparent. Quantitatively, there is no clear difference between the passive and active models of cognitive reserve, as both of them rely on the physiologic basis of biological neural networks in the brain. This provides an opportunity to evaluate the underlying mechanisms that contribute to cognitive reserve across various neurological conditions computationally. 

To avoid any potential confusion between these terms referring to different types of resilience and to avoid any inadvertent conflation between artificial and human brain networks, in the remainder of this paper we will refer to the phenomenon of resilience to damage that we observe in ANNs specifically as ``artificial neural reserve'' and use the terms ``cognitive/brain reserve'' to refer exclusively to human brain networks. 

\subsection{Probing the Neural Network}

The ablation of connections in ANNs is also referred to as probing in NLMs. This is a growing field aimed at understanding the inner workings of large-scale transformer-based NLMs by probing the mechanism (i.e., attention weights, hidden states) to better understand the linguistic structure and representations encoded by such models. Similarly to the early findings of \citet{lukashin1994overlapping} and \citet{lukashin1994neural}, more recent work on transformers (i.e., \citet{NEURIPS2019_2c601ad9, prasanna-etal-2020-bert, NEURIPS2020_eae15aab}) demonstrates that a large percentage of attention heads or sub-modules can be removed at inference time without significantly impacting performance.

\subsection{Linguistic Anomalies in AD}

\begin{figure}[htbp]
\centering
\includegraphics[width=.8\columnwidth]{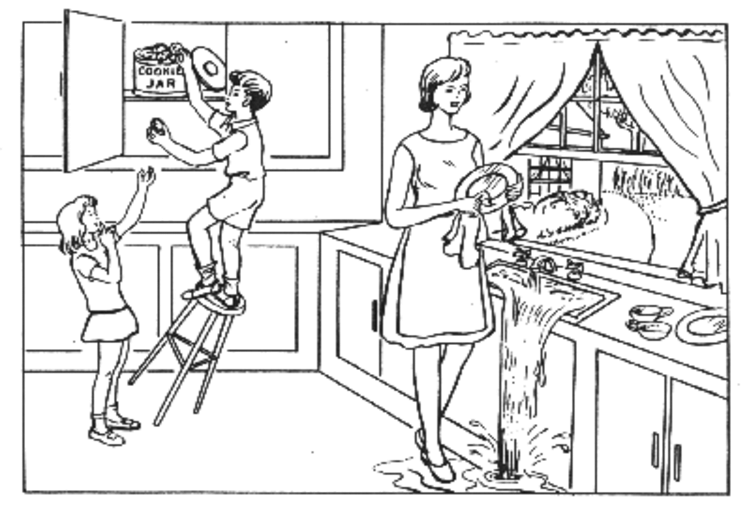}
\caption{The ``Cookie Theft'' picture description stimuli.}
\label{fig:cookie}
\end{figure}

AD is a neurodegerative disease, and progressively worsening linguistic deficits often accompany its progression \citep{kempler2008language, altmann2008effects}. A widely-used diagnostic task to capture such linguistic anomalies is the ``Cookie Theft'' picture description task from the Boston Diagnostic Aphasia Examination \citep{goodglass1983boston}. In this task, participants are asked to describe everything they see going on in Figure~\ref{fig:cookie}. Previous studies have demonstrated that dementia patients tend to overuse pronouns \citep{ALMOR1999202} and tend to perseverate \citep{HIER1985117} when describing the ``Cookie Theft'' picture.

There is a rich body of evidence that supervised machine learning and deep learning methods can learn to distinguish the subtle linguistic characteristics between healthy individuals and people with dementia. However, such models present a danger of overfitting, and hinder interpretability of model predictions, which are both critical concerns for clinical artificial intelligence (AI) applications \citep{graham2020artificial}. Alternatively, PPL is an easily interpretable measure used to evaluate model performance. With dementia, the difference of the \textit{paired-perplexity} paradigm from a ``healthy control'' NLM and a ``dementia'' NLM provides a diagnostically useful summary value that distinguishes language samples produced by dementia patients \citep{fritsch2019automatic, cohen-pakhomov-2020-tale}. Prior work \citep{li-etal-2022-gpt} has shown that the difference of PPLs from a pre-trained GPT-2 paired with an artiﬁcially degraded version of itself approximates SOTA classification performance \textit{without} requiring a data set from dementia patients of comparable size to its comprehensive training data. However, this approach requires evaluating thousands of masking patterns in order to investigate the effects of masking various combinations of attention heads exhaustively. In the current work we obviate this requirement for extensive experimentation by using targeted masking (guided by the changes in gradients during training) of two subsets of attention heads that are a) \textit{most} ``important'', and \textit{least} ``important'' with respect to the content of the ``Cookie Theft'' picture. We show that the resulting masked models can effectively identify transcripts from dementia patients with \textit{significantly fewer} trainable parameters while exhibiting comparable classification performance to previous studies \citep{li-etal-2022-gpt, taghibeyglou2024context}.

\section{Methods}

\subsection{Data}

We use two publicly available datasets that contain responses to the ``Cookie Theft'' picture description task: a) AD Recognition through Spontaneous Speech (ADReSS) Challenge\footnote{\url{https://dementia.talkbank.org/ADReSS-2020/}} \citep{luz20_interspeech}, and b) the Wisconsin Longitudinal Study (WLS)\footnote{\url{https://dementia.talkbank.org/access/English/WLS.html}} \citep{herd2014cohort}. Table~\ref{tab:data} shows basic characteristics of datasets used in this study. ADReSS is a subset of the Pitt corpus \citep{becker1994natural} designed to address the absence of a standardized train/test split in prior work. It is specifically matched on age and gender to reduce potential confounding effects. The WLS is a longitudinal study of 694 men and 675 women who graduated from Wisconsin high schools in 1957. The participants were interviewed up to 6 times between 1957 and 2011. The ``Cookie Theft'' picture description task was administered in the later round of interviews. In particular, we restricted the original WLS dataset to a total of 102 participants who a) agreed to participate in the ``Cookie Theft'' picture description task, and b) had either a clinical diagnosis of dementia or were deemed healthy in follow-up interviews conducted in 2020. This information was obtained through phone interviews and assessments by advanced practice providers. Subsequently, the collected data was presented to a panel of clinicians to obtain the diagnosis.

\begin{table}[htbp]
\centering
\resizebox{\columnwidth}{!}{%
\begin{tabular}{@{}cc|cc@{}}
\toprule
\multicolumn{2}{c|}{\multirow{2}{*}{\textbf{Dataset}}}    & \multicolumn{1}{c|}{\textbf{Dementia}} & \textbf{Healthy Controls} \\ \cmidrule(l){3-4} 
\multicolumn{2}{l|}{}  & \multicolumn{2}{c}{\# of participants ($n$)}    \\ \midrule
\multicolumn{1}{c|}{\multirow{3}{*}{ADReSS}} & Train & \multicolumn{1}{c|}{54} & 54 \\ \cmidrule(l){2-4} 
\multicolumn{1}{l|}{} & Test & \multicolumn{1}{c|}{24} & 24 \\ \cmidrule(l){2-4} 
\multicolumn{1}{l|}{}  & Total & \multicolumn{1}{c|}{78} &78  \\ \midrule
\multicolumn{2}{c|}{WLS} & \multicolumn{1}{c|}{29} & 73\\ \bottomrule
\end{tabular}%
}
\caption{The characteristics of ADReSS and WLS.}
\label{tab:data}
\end{table}

We perform verbatim transcripts pre-processing using TRESTLE (\textbf{T}oolkit for \textbf{R}eproducible \textbf{E}xecution of \textbf{S}peech \textbf{T}ext and \textbf{L}anguage \textbf{E}xperiments) \citep{li2023trestle} by removing utterances that do not belong to the participants, unintelligible words, and speech and non-speech artifacts event descriptions (i.e., ``laughs'', ``clear throat'').

\subsection{Modeling and Evaluation}

We follow a similar masking strategy to that proposed by \citet{NEURIPS2019_2c601ad9} to mask attention heads of the GPT-2 small, medium, large, and XL models via the rank of their importance to the task. We focus on the GPT-2 family models to minimize the variability that would result from multiple modeling architectures. The task-importance of attention heads in each model is determined by the gradient changes during the fine-tuning for subsequent word prediction task using transcripts of ``Cookie Theft'' picture descriptions in the training portion of the ADReSS dataset. Intuitively, if the gradient change of an attention head is large, this attention head is likely important with respect to predicting the language to which the model is being fine-tuned, and vice versa. 

In contrast to the approach by \citet{NEURIPS2019_2c601ad9}, which prunes the \textit{least} important attention heads during testing, we anticipate that the \textit{most} important attention heads are those relevant for predicting the text of the ``Cookie Theft'' task. This idea is supported by \citet{doi:10.1044/jshd.4501.27} and \citet{ berube2019stealing}, who found that the number of content units represented -- a measure of how much relevant information is conveyed in the description -- is sensitive to linguistic deficits often observed in individuals with neurodegenerative disease. However, we also reason that the \textit{least} important attention heads may represent subtle differences in linguistic structure and representations that may distinguish between dementia patients and healthy controls. We also test the possibility that the semantic impairment observed in AD \citep{HUFF1986235, 10.1093/brain/124.8.1522, HODGES1995441} could be potentially simulated by masking a certain share of the columns in the pre-trained NLMs' token embedding matrix, where each column contributes to the representation of the meaning of each token in the model's vocabulary. Thus, masking columns in the embedding matrix leads to degrading the representation of \textit{all} vocabulary items vs. degrading or deleting specific tokens from the otherwise intact vocabulary by operating on the rows of the embedding matrix. 

Following these considerations, we design the masking strategies as follows: a) we fine-tune each of the GPT-2 models with a language model head layer as the top layer on the ADReSS training set to get the corresponding ranking of importance for each of the attention heads; b) we iteratively mask a small share ($n$\%) of ranked attention heads \textit{bidirectionally}, which consists of the $\frac{n}{2}$\% \textit{most} important attention heads and the $\frac{n}{2}$\% \textit{least} important attention heads, then gradually increase the percentage of attention heads for masking, and c) we iteratively mask columns of the word embedding matrix in reverse order, moving from right to left, and gradually increase the percentage of word embedding columns for masking\footnote{All experiments in this study are done with HuggingFace's transformers package \citep{wolf-etal-2020-transformers} on one A100 GPU.}.


We examine the artificial neural reserve hypothesis using two evaluation approaches. The first approach consists of simply estimating the PPL of the progressively degraded NLMs based on healthy individuals' transcripts from an independent dataset containing the same type of picture descriptions as the dataset that was used to rank attention heads by their importance to the dementia classification task. We use the WLS dataset and select only those WLS participants that remained cognitively healthy over the entire study period as the independently collected dataset for log PPL estimation. Using this approach, in addition to masking attention heads, we also experiment with masking model weights in the token embedding matrix to see if any observed effects are specific to the attention mechanism. 

The second approach consists of evaluating the classification performance of ablated/degraded models paired with the original versions of the same GPT-2 model using the paired-perplexity paradigm \citep{fritsch2019automatic, cohen-pakhomov-2020-tale, li-etal-2022-gpt}. These evaluations are conducted on the testing portion of the ADReSS dataset, with accuracy (ACC) and area under the receiver-operator characteristic (ROC) curve (AUC) as the evaluation metrics. Specifically, for the paired-perplexity paradigm, we estimate the ratio of PPLs $\frac{PPL_{control}}{PPL_{dementia}}$ of each transcript from the test set. The ACC measure is calculated as accuracy at the equal error rate (EER), where the false acceptance rate is equal to false rejection rate on the ROC curve. The intuition behind this approach is based on the expectation that successful masking of a portion of attention heads in a pre-trained NLM will result in the NLM exhibiting dementia-like behavior, which would in turn result in high AUC and ACC values of the paired-perplexity classification.

\section{Results}
\subsection{Effects of Masking on Perplexity}

\begin{figure*}[h]
\centering
\begin{subfigure}{\columnwidth} 
  \centering
  \includegraphics[width=0.8\textwidth]{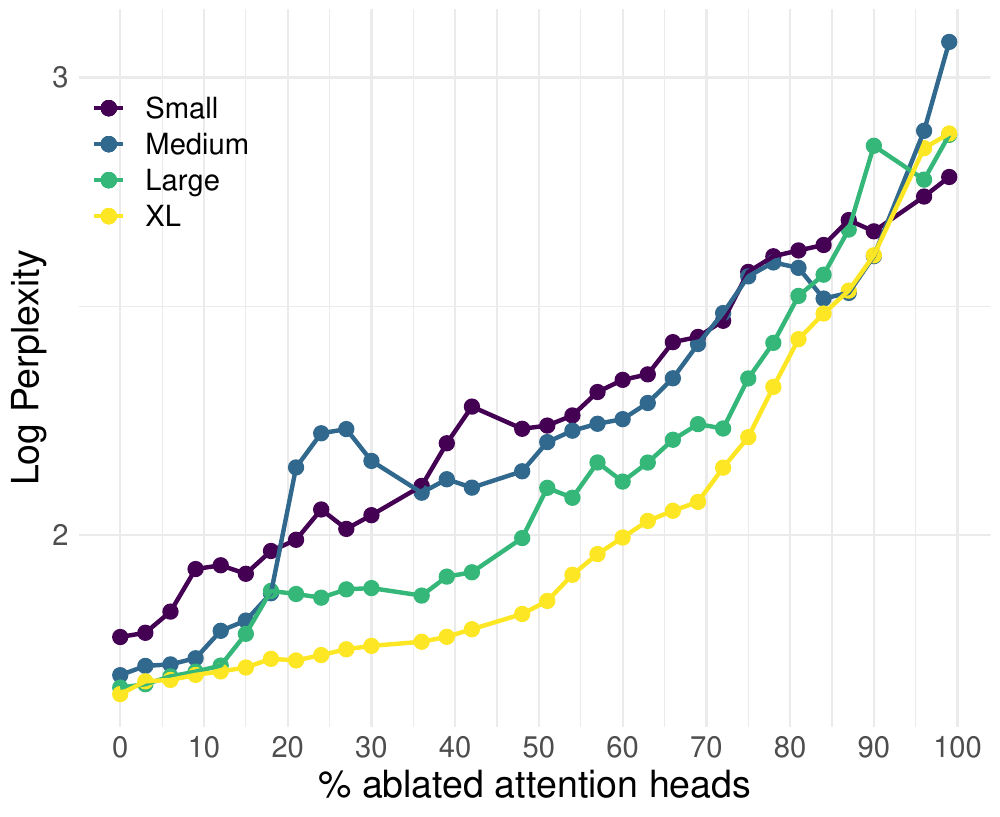} 
  \caption{GPT-2s with masked attention heads.}
  \label{fig:wls-attn}
\end{subfigure}
\vspace{2ex}
\begin{subfigure}{\columnwidth} 
  \centering
  \includegraphics[width=0.8\textwidth]{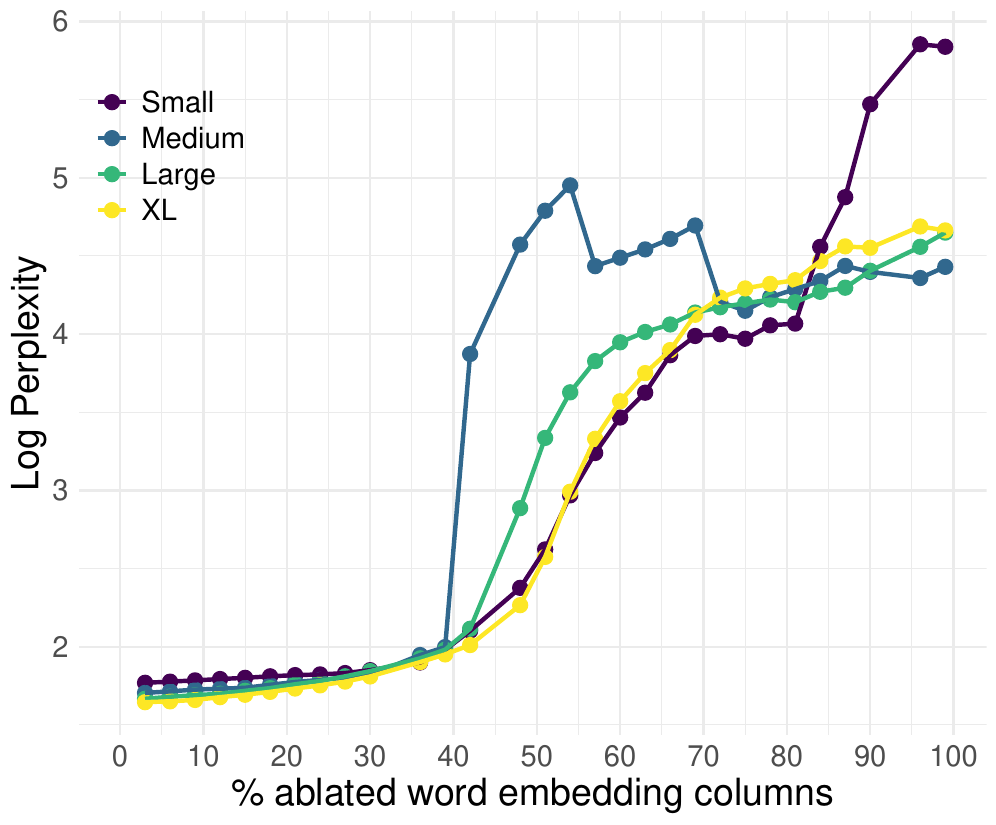}
  \caption{GPT-2s with masked word embedding matrix.}
  \label{fig:wls-wte}
\end{subfigure}
\caption{Changes in model log PPL as a function of the proportion of masked attention heads across GPT-2 models of various sizes. Note: the curves in panel (a) show that GPT-2 XL model has the most non-linear/concave shape indicating that the model starts to degrade rapidly only after masking of about 50\% of its attention heads, followed by the curve for the GPT-2 large model. The smaller GPT-2 models begin to degrade with proportionally less masking, and exhibit a monotonic relationship between the magnitude of attention heads masking and model performance. The curves in panel (b) show almost completely preserved model performance without differences between models up to the point at which 40\% - 50\% of the columns in their embedding matrices have been masked. After that point, the performance of all models collapses ``catastrophically''  }
\label{fig:wls-eval}
\end{figure*}

As illustrated in Figure~\ref{fig:wls-attn}, the predictive ability of smaller GPT-2 models degrades linearly with the degree of damage inflicted on the attention mechanism by masking progressively larger proportion of attention heads. The predictive ability of the larger GPT-2 models, on the other hand, degraded in a non-linear fashion where increases in log PPL were relatively flat up to 40-50\% of the attention heads being masked and then began to increase exponentially. Fitting the GPT-2 small, medium, large and XL model log PPL to a linear regression line resulted in $r^2$ goodness-of-fit values of 0.99, 0.89, 0.91 and 0.83, respectively, whereas fitting to an exponential regression line failed to converge for the small and medium models and yielded $r^2$ values of 0.97 and 0.99 for the large and XL models, respectively. The results of Dunn's test further confirmed our observations, showing that the differences between log PPLs estimated by GPT-2 small and GPT-2 XL (adjusted p-value $<$ 0.01), and GPT-2 medium and GPT-2 XL (adjusted p-value $<$ 0.05) when masking attention heads are statistically significant. In contrast, \textit{all} combinations of log PPLs were not significantly different from each other for all GPT-2 models when masking the word embedding matrix (adjusted p-value $>$ 0.05). 

Compared to masking attention heads, with GPT-2 small, medium, large and XL model we needed to mask 93\% (714 out of 768), 66\% (675 out of 1024), 87\% (1113 out of 1280), and 66\% (1050 out of 1600) columns in the word embedding matrix to achieve ACCs of 0.75, 0.85, 0.79, and 0.81 respectively, on the ADReSS test set. Figure~\ref{fig:wls-wte} can further support this claim, as estimated log PPLs of masking word embedding matrix show no significant statistical differences across various GPT-2 models.

\begin{figure*}[h]
\begin{subfigure}{.5\textwidth}
  \centering
  \includegraphics[width=.8\columnwidth]{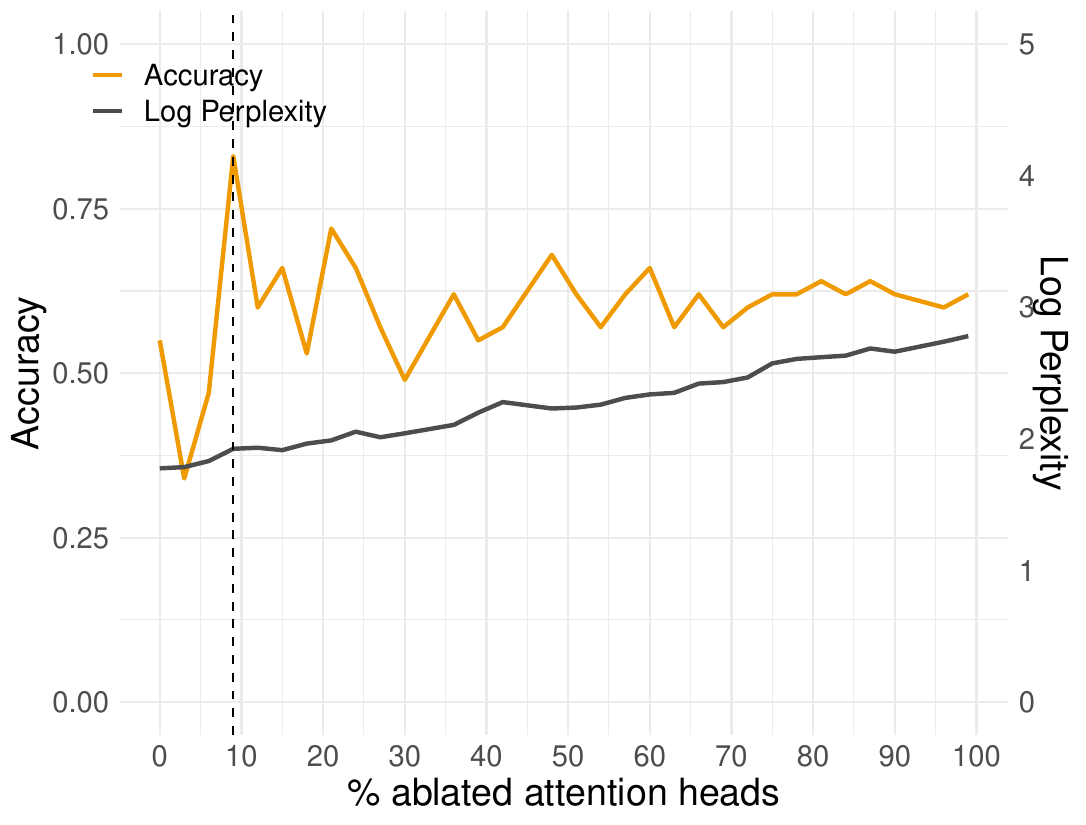}  
  \caption{GPT-2 small}
  \label{fig:gpt2-attn}
\end{subfigure}
\begin{subfigure}{.5\textwidth}
  \centering
  \includegraphics[width=.8\columnwidth]{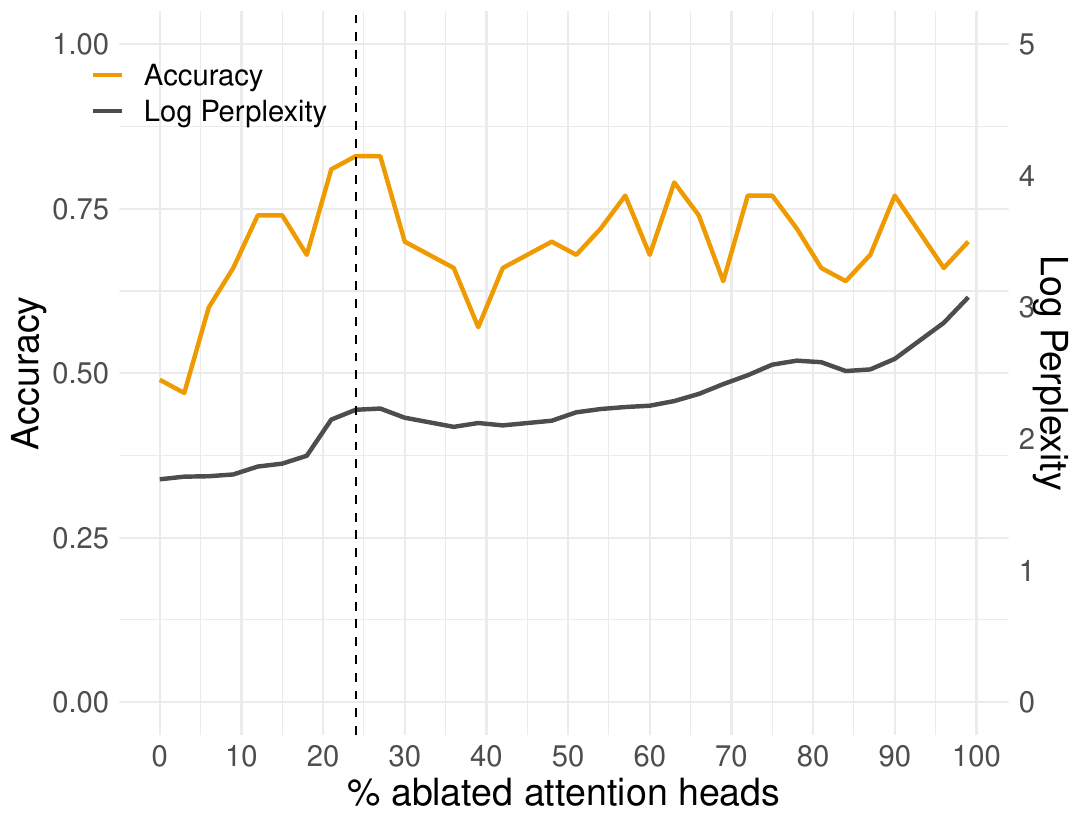}
  \caption{GPT-2 medium}
  \label{fig:gpt2-medium-attn}
\end{subfigure}
\newline
\begin{subfigure}{.5\textwidth}
  \centering
  \includegraphics[width=.8\columnwidth]{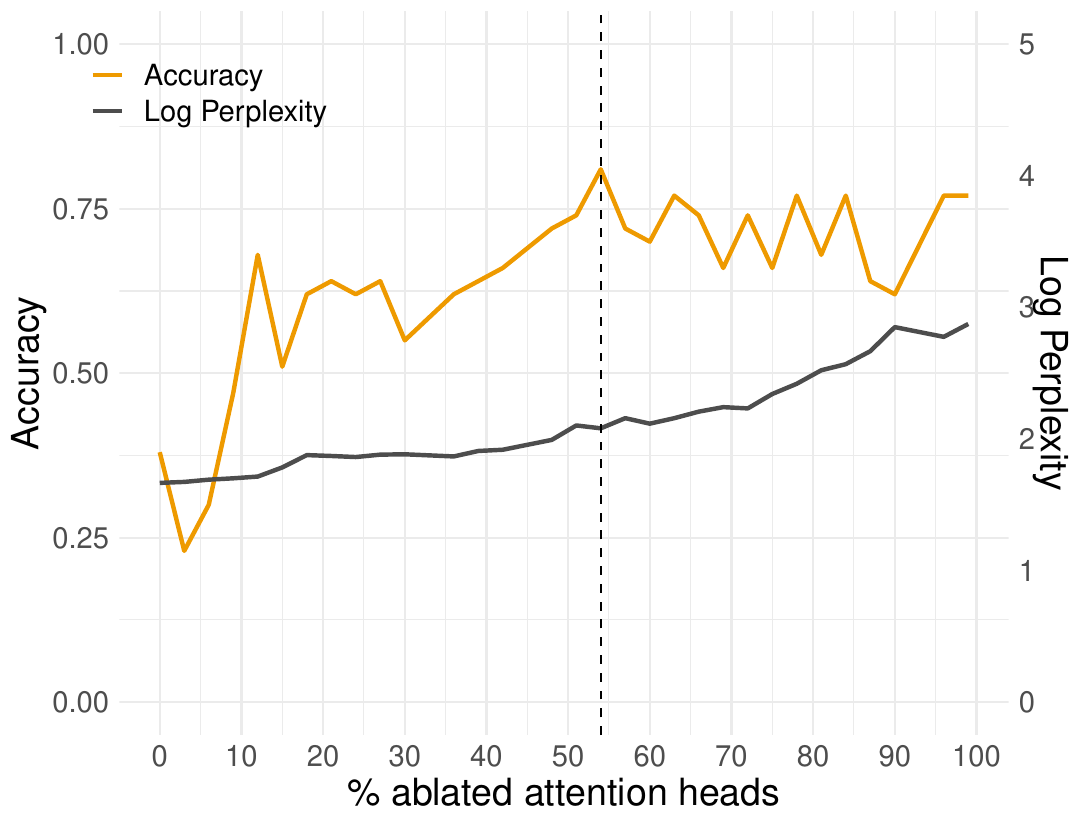}
  \caption{GPT-2 large}
  \label{fig:gpt2-large-attn}
\end{subfigure}
\begin{subfigure}{.5\textwidth}
  \centering
  \includegraphics[width=.8\columnwidth]{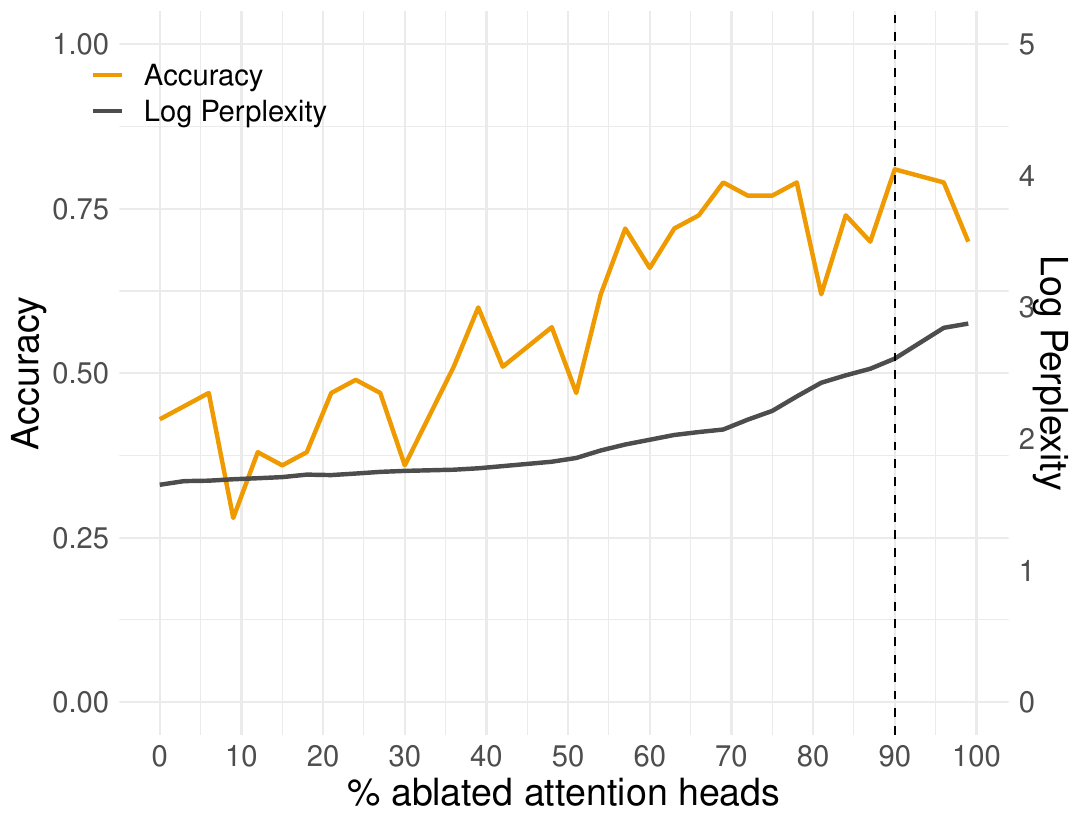}  
  \caption{GPT-2 XL}
  \label{fig:gpt2-xl-attn}
\end{subfigure}
\caption{Comparison of GPT-2 models with masked attention heads on paired-perplexity classification performance. The left y-axis denotes classification performance using both masked and unmasked GPT-2 models on the ADReSS test set. The right y-axis indicates log PPL estimated from transcripts of WLS healthy individuals. The x-axis represents the percentage of attention heads getting masked. The vertical dashed line indicates the best-performing masking pattern, achieving the highest ACC.}
\label{fig:mask-attn}
\end{figure*}

\subsection{Effects of Masking on Dementia Classification}

\begin{table*}[h]
\centering
\begin{tabular}{@{}p{2.4cm}|c|c|c|c@{}}
\toprule
\textbf{Model} & \textbf{GPT-2 small} & \textbf{GPT-2 medium} & \textbf{GPT-2 large} & \textbf{GPT-2 XL} \\ \midrule
\RaggedRight \# of parameters & 124M & 355M & 774M & 1.5B \\ \midrule
\RaggedRight \# of masked attention heads & 12 & 92 & 388 & 1080 \\ \midrule
\RaggedRight \% of masked attention heads & 9 & 24 & 54 & 90 \\ \midrule
ACC & 0.83 & 0.83 & 0.81 & 0.81 \\ \midrule
AUC & 0.86 & 0.85 & 0.80 & 0.82 \\ \bottomrule
\end{tabular}%
\caption{Classification performance of the paired-perplexity approach based on pre-trained and masked GPT-2 models on the ADReSS test set.}
\label{tab:res}
\end{table*}

As shown in Table~\ref{tab:res}, impairing 9\% of attention heads ($n$=12) of the GPT-2 small model (the ``dementia'' model) achieved an ACC of 0.83 and AUC of 0.86 when paired with the original unmasked version of itself (the ``control'' model) on the ADReSS test set. This is comparable to the prior work \citep{li-etal-2022-gpt} (ACC = 0.85, AUC = 0.89) but the masking approach uses \textit{significantly fewer} masked parameters. Our results also show that a larger share of attention heads in the larger models must be masked to approximate a ``dementia'' model with the same level of performance in the paired-perplexity classification than with smaller models. Notably, masking of the word embedding matrix did not result in comparable observations. As anticipated by the results shown in Figure~\ref{fig:wls-wte}, with GPT-2 models we needed to mask a majority portion (e.g., $>$ 50\%) of the word embedding matrix to obtain similar level of classification performance regardless of model size.\footnote{The importance of attention heads for each model can be found in Table~\ref{tab:gpt2-rank}, Table~\ref{tab:gpt2-medium-rank}, Table~\ref{tab:gpt2-large-rank}, and Table~\ref{tab:gpt2-xl-rank} in the Appendix~\ref{sec:appendix}.}

As illustrated in Figure~\ref{fig:mask-attn}, we observed that once the best-performing masking pattern, marked by the highest ACC, was reached, the classification performance of all GPT-2 models started to fluctuate. However, this observation did not occur with the word embedding matrix masking. As illustrated in Figure~\ref{fig:mask-wte} in Appendix~\ref{sec:appendix}, the classification performance exhibited fluctuations prior to the emergence of the best-performing masking pattern, indicating that masking the columns of the word embedding matrix has less impact on identifying the signs of cognitive impairment from text as it probably does not result in a good dementia-like model for the paired-perplexity classification task.

\section{Discussion}

The results of experiments presented in this paper suggest that the notion of cognitive reserve in the brain may have an analogue in transformer-based ANNs that is localized to the attention mechanism. Recent neuroscientific evidence shows that NLMs' PPL is predictive of human behavioral responses and and neural responses in functional MRI studies \citep{doi:10.1073/pnas.2105646118, hosseini2024artificial}. Based on this evidence, we interpret our findings of the differences in log PPL changes as a result of masking attention heads in NLMs of variable size as at least suggestive that the resilience to damage is non-linear to the number of attention heads in NLMs. In other words, it takes disproportionately more masking to damage larger NLMs to elicit the same level of degradation in performance, as compared with smaller NLMs. Furthermore, the dissociation in performance as a result of damaging attention heads vs. the token embedding weights suggests that the NLM's artificial neural reserve effects are localized to the attention mechanism.

Our results also suggest that masking attention heads within the paired-perplexity paradigm using the ratio of unmasked (``control'') and masked (``dementia'') pre-trained GPT-2 models results in good classification performance \textit{without} requiring a corresponding large dataset produced by dementia patients and extensive parameter tuning. This can be achieved with as little as masking only 9\% of attention heads of a pre-trained GPT-2 small model. 



In contrast to previous studies, which typically involved purging attention heads determined to be the \textit{least} important, our bidirectional masking method adds supporting evidence of content units \citep{doi:10.1044/jshd.4501.27, berube2019stealing}, suggesting the importance of these contextual features in addition to the predominant emphasis on linguistic structure and representation modeling in previous research (e.g., \citet{orimaye-etal-2014-learning}, \citet{fraser2016linguistic}). The results of bidirectional masking also offers an interpretable explanation for transfer learning's remarkable performance using pre-trained NLMs. It suggests that during fine-tuning, pre-trained NLMs use a combination of both task-specific (the \textit{most} important) and task-agnostic (the \textit{least} important) heads to achieve remarkable performance on various downstream tasks. Those task-agnostic attention heads may play an important role in transfer learning. This also may explain why distilled NLMs in which the ``nonvolitional cues'' that fall outside of common NLP benchmarks are purged during the distillation, generalize less-than-ideally to other types of data produced by individuals with dementia \citep{li-etal-2022-gpt}. With larger models, there are considerably more attention heads that can serve as those ``nonvolitional cues,'' helping a larger NLM perform better \citep{agbavor2022predicting}.

As the columns of the token embedding matrix in a pre-trained NLMs represent the global semantics of tokens in the vocabulary, the observations that the best-performing masking pattern appears in the later stage of the token embedding matrix masking are consistent with previously published findings that semantic impairment often occurs in the later stage (i.e., moderate) of the disease \citep{HUFF1986235, 10.1093/brain/124.8.1522, HODGES1995441}. As illustrated in Figure~\ref{fig:mask-wte}, when masking the later 66\% columns (675 out of 1024) of the word embedding matrix, the paired unmasked and masked GPT-2 medium achieves an ACC of 0.85 on the ADReSS test set. This finding is consistent with a previous work \citep{hewitt-manning-2019-structural}, suggesting that some syntactic information is embedded  implicitly in the word embedding matrix. This also provide an empirical support of our findings that masking word embedding matrix of a pre-trained NLM can provide some degree of discriminating effect on this downstream task. However, masking the word embedding matrix is far less effective than masking attention heads to simulate dementia-related cognitive impairment.  

Our results suggest that similar mechanisms of resilience may exist in both human cognition and computational models. This could lead to more nuanced strategies in response to develop early screening tools for the delayed onset of cognitive impairment in the population with high risk. Our study holds promise for deploying such early-screening method in resource-constrained clinical settings to improve early intervention and patient management for AD. 



\section{Conclusion}

We presented experimental findings suggesting the presence of  artificial neural reserve in transformer-based NLMs, analogous to the concepts of brain/cognitive reserve in studies of human cognition.  In addition, we introduced a novel bidirectional attention head ablation method that enables using unmasked and masked GPT-2 models in the paired-perplexity paradigm for detecting linguistic anomalies with significantly less parameter masking or fitting. 


\section*{Acknowledgement}
Research reported in this publication was supported by the National Library of Medicine of the National Institutes of Health under award number R01 LM014056-02S1, and the National Institute on Aging of the National institutes of Health under award number AG069792.

\section*{Limitations}
The work presented here has several limitations. First, the size of datasets used in this study is relatively small compared to datasets typically analyzed in the open-domain NLP tasks, therefore the results may not be readily generalizable. Second, all datasets used in our study are in American English, and many participants of these two studies are representative of White, non-Hispanic American men and women located at the north part of the United States, which certainly limit their applicability to other languages. Third, while we propose that the findings presented in this paper may be interpreted as an analogue of the notions of cognitive or brain reserve, we do not suggest that GPT-2 models are accurate models of the human brain. Rather, our interpretation of these findings is that experimenting with masking of attention heads in models of various sizes and architectures may be useful in helping us understand cognitive processes that take place in the human brain. The observed effects of attention masking on the model's performance and behavior, while suggestive of an analog to cognitive reserve in the human brain, should not establish a direct causal link to human cognitive processes. Additionally, the ranking of attention heads by their relative importance is specific to the ADReSS dataset as it was derived in the training portion of the dataset and may not readily generalize to other datasets and types of data. Lastly, in this paper we did not address the distinction between the notions of cognitive and brain reserves. It would be important to investigate in future work if NLMs of the same size and architecture but different quantities and quality of the training data (i.e., as a simulation of educational attainment) exhibit differential resilience to damage independently of the effects observed in models of variable size.


\bibliography{anthology,custom}

\begin{thebibliography}{50}
\expandafter\ifx\csname natexlab\endcsname\relax\def\natexlab#1{#1}\fi

\bibitem[{Agbavor and Liang(2022)}]{agbavor2022predicting}
Felix Agbavor and Hualou Liang. 2022.
\newblock \href {https://doi.org/https://doi.org/10.1371/journal.pdig.0000168}
  {Predicting dementia from spontaneous speech using large language models}.
\newblock \emph{PLOS Digital Health}, 1(12):e0000168.

\bibitem[{Almor et~al.(1999)Almor, Kempler, MacDonald, Andersen, and
  Tyler}]{ALMOR1999202}
Amit Almor, Daniel Kempler, Maryellen~C MacDonald, Elaine~S Andersen, and
  Lorraine~K Tyler. 1999.
\newblock \href {https://doi.org/https://doi.org/10.1006/brln.1999.2055} {Why
  do alzheimer patients have difficulty with pronouns? working memory,
  semantics, and reference in comprehension and production in alzheimer's
  disease}.
\newblock \emph{Brain and Language}, 67(3):202--227.

\bibitem[{Altmann and McClung(2008)}]{altmann2008effects}
Lori~JP Altmann and Jill~S McClung. 2008.
\newblock \href {https://doi.org/10.1055/s-2008-1061622} {Effects of semantic
  impairment on language use in alzheimer's disease}.
\newblock In \emph{Seminars in speech and language}, volume~29, pages 018--031.
  {\copyright} Thieme Medical Publishers.

\bibitem[{Becker et~al.(1994)Becker, Boiler, Lopez, Saxton, and
  McGonigle}]{becker1994natural}
James~T Becker, Fran{\c{c}}ois Boiler, Oscar~L Lopez, Judith Saxton, and
  Karen~L McGonigle. 1994.
\newblock The natural history of alzheimer's disease: description of study
  cohort and accuracy of diagnosis.
\newblock \emph{Archives of neurology}, 51(6):585--594.

\bibitem[{Berube et~al.(2019)Berube, Nonnemacher, Demsky, Glenn, Saxena,
  Wright, Tippett, and Hillis}]{berube2019stealing}
Shauna Berube, Jodi Nonnemacher, Cornelia Demsky, Shenly Glenn, Sadhvi Saxena,
  Amy Wright, Donna~C Tippett, and Argye~E Hillis. 2019.
\newblock \href {https://doi.org/10.1044/2018_AJSLP-17-0131} {Stealing cookies
  in the twenty-first century: Measures of spoken narrative in healthy versus
  speakers with aphasia}.
\newblock \emph{American journal of speech-language pathology},
  28(1S):321--329.

\bibitem[{Bradford et~al.(2009)Bradford, Kunik, Schulz, Williams, and
  Singh}]{bradford2009missed}
Andrea Bradford, Mark~E Kunik, Paul Schulz, Susan~P Williams, and Hardeep
  Singh. 2009.
\newblock \href {https://doi.org/10.1097/WAD.0b013e3181a6bebc} {Missed and
  delayed diagnosis of dementia in primary care: prevalence and contributing
  factors}.
\newblock \emph{Alzheimer disease and associated disorders}, 23(4):306.

\bibitem[{Cohen and Pakhomov(2020)}]{cohen-pakhomov-2020-tale}
Trevor Cohen and Serguei Pakhomov. 2020.
\newblock \href {https://doi.org/10.18653/v1/2020.acl-main.176} {A tale of two
  perplexities: Sensitivity of neural language models to lexical retrieval
  deficits in dementia of the {A}lzheimer{'}s type}.
\newblock In \emph{Proceedings of the 58th Annual Meeting of the Association
  for Computational Linguistics}, pages 1946--1957, Online. Association for
  Computational Linguistics.

\bibitem[{Esiri et~al.(2001)Esiri, Matthews, Brayne, Ince, Matthews, Xuereb,
  Broome, McKenzie, Rossi, McKeith et~al.}]{esiri2001pathological}
MM~Esiri, F~Matthews, C~Brayne, PG~Ince, FE~Matthews, JH~Xuereb, JC~Broome,
  J~McKenzie, M~Rossi, IG~McKeith, et~al. 2001.
\newblock \href {https://doi.org/10.1016/S0140-6736(00)03589-3} {Pathological
  correlates of late-onset dementia in a multicentre, community-based
  population in england and wales}.
\newblock \emph{Lancet}.

\bibitem[{Fraser et~al.(2016)Fraser, Meltzer, and
  Rudzicz}]{fraser2016linguistic}
Kathleen~C Fraser, Jed~A Meltzer, and Frank Rudzicz. 2016.
\newblock \href {https://doi.org/10.3233/JAD-150520} {Linguistic features
  identify alzheimer’s disease in narrative speech}.
\newblock \emph{Journal of Alzheimer's Disease}, 49(2):407--422.

\bibitem[{Fritsch et~al.(2019)Fritsch, Wankerl, and
  N{\"o}th}]{fritsch2019automatic}
Julian Fritsch, Sebastian Wankerl, and Elmar N{\"o}th. 2019.
\newblock \href {https://doi.org/10.1109/ICASSP.2019.8682690} {Automatic
  diagnosis of alzheimer’s disease using neural network language models}.
\newblock In \emph{ICASSP 2019-2019 IEEE International Conference on Acoustics,
  Speech and Signal Processing (ICASSP)}, pages 5841--5845. IEEE.

\bibitem[{Giffard et~al.(2001)Giffard, Desgranges, Nore-Mary, Lalevée, de~la
  Sayette, Pasquier, and Eustache}]{10.1093/brain/124.8.1522}
Bénédicte Giffard, Béatrice Desgranges, Florence Nore-Mary, Catherine
  Lalevée, Vincent de~la Sayette, Florence Pasquier, and Francis Eustache.
  2001.
\newblock \href {https://doi.org/10.1093/brain/124.8.1522} {{The nature of
  semantic memory deficits in Alzheimer's disease: New insights from
  hyperpriming effects}}.
\newblock \emph{Brain}, 124(8):1522--1532.

\bibitem[{Giles et~al.(1996)Giles, Patterson, and
  Hodges}]{doi:10.1080/02687039608248419}
Elaine Giles, Karalyn Patterson, and John~R Hodges. 1996.
\newblock \href {https://doi.org/10.1080/02687039608248419} {Performance on the
  boston cookie theft picture description task in patients with early dementia
  of the alzheimer's type: Missing information}.
\newblock \emph{Aphasiology}, 10(4):395--408.

\bibitem[{Goodglass and Kaplan(1983)}]{goodglass1983boston}
Harold Goodglass and Edith Kaplan. 1983.
\newblock \emph{Boston diagnostic aphasia examination booklet}.
\newblock Lea \& Febiger.

\bibitem[{Graham et~al.(2020)Graham, Lee, Jeste, Van~Patten, Twamley, Nebeker,
  Yamada, Kim, and Depp}]{graham2020artificial}
Sarah~A Graham, Ellen~E Lee, Dilip~V Jeste, Ryan Van~Patten, Elizabeth~W
  Twamley, Camille Nebeker, Yasunori Yamada, Ho-Cheol Kim, and Colin~A Depp.
  2020.
\newblock \href
  {https://doi.org/https://doi.org/10.1016/j.psychres.2019.112732} {Artificial
  intelligence approaches to predicting and detecting cognitive decline in
  older adults: A conceptual review}.
\newblock \emph{Psychiatry research}, 284:112732.

\bibitem[{Herd et~al.(2014)Herd, Carr, and Roan}]{herd2014cohort}
Pamela Herd, Deborah Carr, and Carol Roan. 2014.
\newblock Cohort profile: Wisconsin longitudinal study (wls).
\newblock \emph{International journal of epidemiology}, 43(1):34--41.

\bibitem[{Hewitt and Manning(2019)}]{hewitt-manning-2019-structural}
John Hewitt and Christopher~D. Manning. 2019.
\newblock \href {https://doi.org/10.18653/v1/N19-1419} {{A} structural probe
  for finding syntax in word representations}.
\newblock In \emph{Proceedings of the 2019 Conference of the North {A}merican
  Chapter of the Association for Computational Linguistics: Human Language
  Technologies, Volume 1 (Long and Short Papers)}, pages 4129--4138,
  Minneapolis, Minnesota. Association for Computational Linguistics.

\bibitem[{Hier et~al.(1985)Hier, Hagenlocker, and Shindler}]{HIER1985117}
Daniel~B. Hier, Karen Hagenlocker, and Andrea~Gellin Shindler. 1985.
\newblock \href {https://doi.org/https://doi.org/10.1016/0093-934X(85)90124-5}
  {Language disintegration in dementia: Effects of etiology and severity}.
\newblock \emph{Brain and Language}, 25(1):117--133.

\bibitem[{Hodges and Patterson(1995)}]{HODGES1995441}
John~R. Hodges and Karalyn Patterson. 1995.
\newblock \href {https://doi.org/https://doi.org/10.1016/0028-3932(94)00127-B}
  {Is semantic memory consistently impaired early in the course of alzheimer's
  disease? neuroanatomical and diagnostic implications}.
\newblock \emph{Neuropsychologia}, 33(4):441--459.

\bibitem[{Hosseini et~al.(2024)Hosseini, Schrimpf, Zhang, Bowman, Zaslavsky,
  and Fedorenko}]{hosseini2024artificial}
Eghbal~A Hosseini, Martin Schrimpf, Yian Zhang, Samuel~R Bowman, Noga
  Zaslavsky, and Evelina Fedorenko. 2024.
\newblock \href {https://doi.org/https://doi.org/10.1162/nol_a_00137}
  {Artificial neural network language models predict human brain responses to
  language even after a developmentally realistic amount of training}.
\newblock \emph{Neurobiology of Language}, pages 1--50.

\bibitem[{Huff et~al.(1986)Huff, Corkin, and Growdon}]{HUFF1986235}
F.Jacob Huff, Suzanne Corkin, and John~H. Growdon. 1986.
\newblock \href {https://doi.org/https://doi.org/10.1016/0093-934X(86)90103-3}
  {Semantic impairment and anomia in alzheimer's disease}.
\newblock \emph{Brain and Language}, 28(2):235--249.

\bibitem[{Iraniparast et~al.(2022)Iraniparast, Shi, Wu, Zeng, Maxwell, Kryscio,
  St~John, SantaCruz, and Tyas}]{iraniparast2022cognitive}
Maryam Iraniparast, Yidan Shi, Ying Wu, Leilei Zeng, Colleen~J Maxwell,
  Richard~J Kryscio, Philip~D St~John, Karen~S SantaCruz, and Suzanne~L Tyas.
  2022.
\newblock \href {https://doi.org/https://doi.org/10.1212/WNL.0000000000200051}
  {Cognitive reserve and mild cognitive impairment: predictors and rates of
  reversion to intact cognition vs progression to dementia}.
\newblock \emph{Neurology}, 98(11):e1114--e1123.

\bibitem[{Kanerva(2009)}]{kanerva2009hyperdimensional}
Pentti Kanerva. 2009.
\newblock \href {https://doi.org/https://doi.org/10.1007/s12559-009-9009-8}
  {Hyperdimensional computing: An introduction to computing in distributed
  representation with high-dimensional random vectors}.
\newblock \emph{Cognitive computation}, 1:139--159.

\bibitem[{Katzman(1993)}]{katzman1993education}
Robert Katzman. 1993.
\newblock \href {https://doi.org/https://doi.org/10.1212/WNL.43.1_Part_1.13}
  {Education and the prevalence of dementia and alzheimer's disease}.
\newblock \emph{Neurology}, 43(1\_part\_1):13--13.

\bibitem[{Katzman et~al.(1988)Katzman, Terry, DeTeresa, Brown, Davies, Fuld,
  Renbing, and Peck}]{katzman1988clinical}
Robert Katzman, Robert Terry, Richard DeTeresa, Theodore Brown, Peter Davies,
  Paula Fuld, Xiong Renbing, and Arthur Peck. 1988.
\newblock \href {https://doi.org/https://doi.org/10.1002/ana.410230206}
  {Clinical, pathological, and neurochemical changes in dementia: a subgroup
  with preserved mental status and numerous neocortical plaques}.
\newblock \emph{Annals of Neurology: Official Journal of the American
  Neurological Association and the Child Neurology Society}, 23(2):138--144.

\bibitem[{Kempler and Goral(2008)}]{kempler2008language}
Daniel Kempler and Mira Goral. 2008.
\newblock \href {https://doi.org/10.1017/S0267190508080045} {Language and
  dementia: Neuropsychological aspects}.
\newblock \emph{Annual review of applied linguistics}, 28:73--90.

\bibitem[{Li et~al.(2022)Li, Knopman, Xu, Cohen, and
  Pakhomov}]{li-etal-2022-gpt}
Changye Li, David Knopman, Weizhe Xu, Trevor Cohen, and Serguei Pakhomov. 2022.
\newblock \href {https://doi.org/10.18653/v1/2022.acl-long.131} {{GPT}-{D}:
  Inducing dementia-related linguistic anomalies by deliberate degradation of
  artificial neural language models}.
\newblock In \emph{Proceedings of the 60th Annual Meeting of the Association
  for Computational Linguistics (Volume 1: Long Papers)}, pages 1866--1877,
  Dublin, Ireland. Association for Computational Linguistics.

\bibitem[{Li et~al.(2023)Li, Xu, Cohen, Michalowski, and
  Pakhomov}]{li2023trestle}
Changye Li, Weizhe Xu, Trevor Cohen, Martin Michalowski, and Serguei Pakhomov.
  2023.
\newblock Trestle: Toolkit for reproducible execution of speech, text and
  language experiments.
\newblock \emph{AMIA Summits on Translational Science Proceedings}, 2023:360.

\bibitem[{Lukashin and Georgopoulos(1994)}]{lukashin1994neural}
Alexander~V Lukashin and Apostolos~P Georgopoulos. 1994.
\newblock \href {https://doi.org/10.1162/neco.1994.6.1.19} {A neural network
  for coding of trajectories by time series of neuronal population vectors}.
\newblock \emph{Neural Computation}, 6(1):19--28.

\bibitem[{Lukashin et~al.(1994)Lukashin, Wilcox, and
  Georgopoulos}]{lukashin1994overlapping}
Alexander~V Lukashin, George~L Wilcox, and Apostolos~P Georgopoulos. 1994.
\newblock \href {https://doi.org/https://doi.org/10.1073/pnas.91.18.8651}
  {Overlapping neural networks for multiple motor engrams.}
\newblock \emph{Proceedings of the National Academy of Sciences},
  91(18):8651--8654.

\bibitem[{Luz et~al.(2020)Luz, Haider, de~la Fuente, Fromm, and
  MacWhinney}]{luz20_interspeech}
Saturnino Luz, Fasih Haider, Sofia de~la Fuente, Davida Fromm, and Brian
  MacWhinney. 2020.
\newblock \href {https://doi.org/10.21437/Interspeech.2020-2571}
  {{Alzheimer’s Dementia Recognition Through Spontaneous Speech: The ADReSS
  Challenge}}.
\newblock In \emph{Proc. Interspeech 2020}, pages 2172--2176.

\bibitem[{Michel et~al.(2019)Michel, Levy, and Neubig}]{NEURIPS2019_2c601ad9}
Paul Michel, Omer Levy, and Graham Neubig. 2019.
\newblock \href
  {https://proceedings.neurips.cc/paper_files/paper/2019/file/2c601ad9d2ff9bc8b282670cdd54f69f-Paper.pdf}
  {Are sixteen heads really better than one?}
\newblock In \emph{Advances in Neural Information Processing Systems},
  volume~32. Curran Associates, Inc.

\bibitem[{Orimaye et~al.(2014)Orimaye, Wong, and
  Golden}]{orimaye-etal-2014-learning}
Sylvester~Olubolu Orimaye, Jojo Sze-Meng Wong, and Karen~Jennifer Golden. 2014.
\newblock \href {https://doi.org/10.3115/v1/W14-3210} {Learning predictive
  linguistic features for {A}lzheimer{'}s disease and related dementias using
  verbal utterances}.
\newblock In \emph{Proceedings of the Workshop on Computational Linguistics and
  Clinical Psychology: From Linguistic Signal to Clinical Reality}, pages
  78--87, Baltimore, Maryland, USA. Association for Computational Linguistics.

\bibitem[{Prasanna et~al.(2020)Prasanna, Rogers, and
  Rumshisky}]{prasanna-etal-2020-bert}
Sai Prasanna, Anna Rogers, and Anna Rumshisky. 2020.
\newblock \href {https://doi.org/10.18653/v1/2020.emnlp-main.259} {{W}hen
  {BERT} {P}lays the {L}ottery, {A}ll {T}ickets {A}re {W}inning}.
\newblock In \emph{Proceedings of the 2020 Conference on Empirical Methods in
  Natural Language Processing (EMNLP)}, pages 3208--3229, Online. Association
  for Computational Linguistics.

\bibitem[{Radford et~al.(2019)Radford, Wu, Child, Luan, Amodei, Sutskever
  et~al.}]{radford2019language}
Alec Radford, Jeffrey Wu, Rewon Child, David Luan, Dario Amodei, Ilya
  Sutskever, et~al. 2019.
\newblock Language models are unsupervised multitask learners.
\newblock \emph{OpenAI blog}, 1(8):9.

\bibitem[{Sanh et~al.(2020)Sanh, Wolf, and Rush}]{NEURIPS2020_eae15aab}
Victor Sanh, Thomas Wolf, and Alexander Rush. 2020.
\newblock \href
  {https://proceedings.neurips.cc/paper_files/paper/2020/file/eae15aabaa768ae4a5993a8a4f4fa6e4-Paper.pdf}
  {Movement pruning: Adaptive sparsity by fine-tuning}.
\newblock In \emph{Advances in Neural Information Processing Systems},
  volume~33, pages 20378--20389. Curran Associates, Inc.

\bibitem[{Satz(1993)}]{satz1993brain}
Paul Satz. 1993.
\newblock \href {https://doi.org/https://doi.org/10.1037/0894-4105.7.3.273}
  {Brain reserve capacity on symptom onset after brain injury: a formulation
  and review of evidence for threshold theory.}
\newblock \emph{Neuropsychology}, 7(3):273.

\bibitem[{Scarmeas and Stern(2003)}]{scarmeas2003cognitive}
Nikolaos Scarmeas and Yaakov Stern. 2003.
\newblock \href
  {https://doi.org/https://doi-org.ezp1.lib.umn.edu/10.1076/jcen.25.5.625.14576}
  {Cognitive reserve and lifestyle}.
\newblock \emph{Journal of clinical and experimental neuropsychology},
  25(5):625--633.

\bibitem[{Scarmeas and Stern(2004)}]{scarmeas2004cognitive}
Nikolaos Scarmeas and Yaakov Stern. 2004.
\newblock \href {https://doi.org/https://doi.org/10.1007/s11910-004-0084-7}
  {Cognitive reserve: implications for diagnosis and prevention of
  alzheimer’s disease}.
\newblock \emph{Current neurology and neuroscience reports}, 4:374--380.

\bibitem[{Schrimpf et~al.(2021)Schrimpf, Blank, Tuckute, Kauf, Hosseini,
  Kanwisher, Tenenbaum, and Fedorenko}]{doi:10.1073/pnas.2105646118}
Martin Schrimpf, Idan~Asher Blank, Greta Tuckute, Carina Kauf, Eghbal~A.
  Hosseini, Nancy Kanwisher, Joshua~B. Tenenbaum, and Evelina Fedorenko. 2021.
\newblock \href {https://doi.org/10.1073/pnas.2105646118} {The neural
  architecture of language: Integrative modeling converges on predictive
  processing}.
\newblock \emph{Proceedings of the National Academy of Sciences},
  118(45):e2105646118.

\bibitem[{Shi et~al.(2023)Shi, Cheung, and Shahamiri}]{SHI2023115538}
Mengke Shi, Gary Cheung, and Seyed~Reza Shahamiri. 2023.
\newblock \href
  {https://doi.org/https://doi.org/10.1016/j.psychres.2023.115538} {Speech and
  language processing with deep learning for dementia diagnosis: A systematic
  review}.
\newblock \emph{Psychiatry Research}, 329:115538.

\bibitem[{Snowdon et~al.(1996)Snowdon, Kemper, Mortimer, Greiner, Wekstein, and
  Markesbery}]{snowdon1996linguistic}
David~A Snowdon, Susan~J Kemper, James~A Mortimer, Lydia~H Greiner, David~R
  Wekstein, and William~R Markesbery. 1996.
\newblock \href {https://doi.org/10.1001/jama.1996.03530310034029} {Linguistic
  ability in early life and cognitive function and alzheimer's disease in late
  life: Findings from the nun study}.
\newblock \emph{Jama}, 275(7):528--532.

\bibitem[{Steffener and Stern(2012)}]{steffener2012exploring}
Jason Steffener and Yaakov Stern. 2012.
\newblock \href {https://doi.org/https://doi.org/10.1016/j.bbadis.2011.09.012}
  {Exploring the neural basis of cognitive reserve in aging}.
\newblock \emph{Biochimica et Biophysica Acta (BBA)-Molecular Basis of
  Disease}, 1822(3):467--473.

\bibitem[{Stern(2002)}]{stern2002cognitive}
Yaakov Stern. 2002.
\newblock \href {https://doi.org/doi:10.1017/S1355617702813248} {What is
  cognitive reserve? theory and research application of the reserve concept}.
\newblock \emph{Journal of the international neuropsychological society},
  8(3):448--460.

\bibitem[{Stern(2009)}]{stern2009cognitive}
Yaakov Stern. 2009.
\newblock \href
  {https://doi.org/https://doi.org/10.1016/j.neuropsychologia.2009.03.004}
  {Cognitive reserve}.
\newblock \emph{Neuropsychologia}, 47(10):2015--2028.

\bibitem[{Stern(2012)}]{stern2012cognitive}
Yaakov Stern. 2012.
\newblock \href {https://doi.org/https://doi.org/10.1016/S1474-4422(12)70191-6}
  {Cognitive reserve in ageing and alzheimer's disease}.
\newblock \emph{The Lancet Neurology}, 11(11):1006--1012.

\bibitem[{Stokes et~al.(2015)Stokes, Combes, and
  Stokes}]{doi.org/10.1111/psyg.12095}
Laura Stokes, Helen Combes, and Graham Stokes. 2015.
\newblock \href {https://doi.org/https://doi.org/10.1111/psyg.12095} {The
  dementia diagnosis: a literature review of information, understanding, and
  attributions}.
\newblock \emph{Psychogeriatrics}, 15(3):218--225.

\bibitem[{TaghiBeyglou and Rudzicz(2024)}]{taghibeyglou2024context}
Behrad TaghiBeyglou and Frank Rudzicz. 2024.
\newblock \href {https://doi.org/https://doi.org/10.1016/j.nlp.2023.100046}
  {Context is not key: Detecting alzheimer’s disease with both classical and
  transformer-based neural language models}.
\newblock \emph{Natural Language Processing Journal}, 6:100046.

\bibitem[{Vaswani et~al.(2017)Vaswani, Shazeer, Parmar, Uszkoreit, Jones,
  Gomez, Kaiser, and Polosukhin}]{NIPS2017_3f5ee243}
Ashish Vaswani, Noam Shazeer, Niki Parmar, Jakob Uszkoreit, Llion Jones,
  Aidan~N Gomez, \L~ukasz Kaiser, and Illia Polosukhin. 2017.
\newblock \href
  {https://proceedings.neurips.cc/paper_files/paper/2017/file/3f5ee243547dee91fbd053c1c4a845aa-Paper.pdf}
  {Attention is all you need}.
\newblock In \emph{Advances in Neural Information Processing Systems},
  volume~30. Curran Associates, Inc.

\bibitem[{Wolf et~al.(2020)Wolf, Debut, Sanh, Chaumond, Delangue, Moi, Cistac,
  Rault, Louf, Funtowicz, Davison, Shleifer, von Platen, Ma, Jernite, Plu, Xu,
  Le~Scao, Gugger, Drame, Lhoest, and Rush}]{wolf-etal-2020-transformers}
Thomas Wolf, Lysandre Debut, Victor Sanh, Julien Chaumond, Clement Delangue,
  Anthony Moi, Pierric Cistac, Tim Rault, Remi Louf, Morgan Funtowicz, Joe
  Davison, Sam Shleifer, Patrick von Platen, Clara Ma, Yacine Jernite, Julien
  Plu, Canwen Xu, Teven Le~Scao, Sylvain Gugger, Mariama Drame, Quentin Lhoest,
  and Alexander Rush. 2020.
\newblock \href {https://doi.org/10.18653/v1/2020.emnlp-demos.6} {Transformers:
  State-of-the-art natural language processing}.
\newblock In \emph{Proceedings of the 2020 Conference on Empirical Methods in
  Natural Language Processing: System Demonstrations}, pages 38--45, Online.
  Association for Computational Linguistics.

\bibitem[{Yorkston and Beukelman(1980)}]{doi:10.1044/jshd.4501.27}
Kathryn~M. Yorkston and David~R. Beukelman. 1980.
\newblock \href {https://doi.org/10.1044/jshd.4501.27} {An analysis of
  connected speech samples of aphasic and normal speakers}.
\newblock \emph{Journal of Speech and Hearing Disorders}, 45(1):27--36.

\end{thebibliography}

\appendix

\section{Appendix}
\label{sec:appendix}


\begin{figure*}[htbp]

\begin{subfigure}{.5\textwidth}
  \centering
  \includegraphics[width=.8\columnwidth]{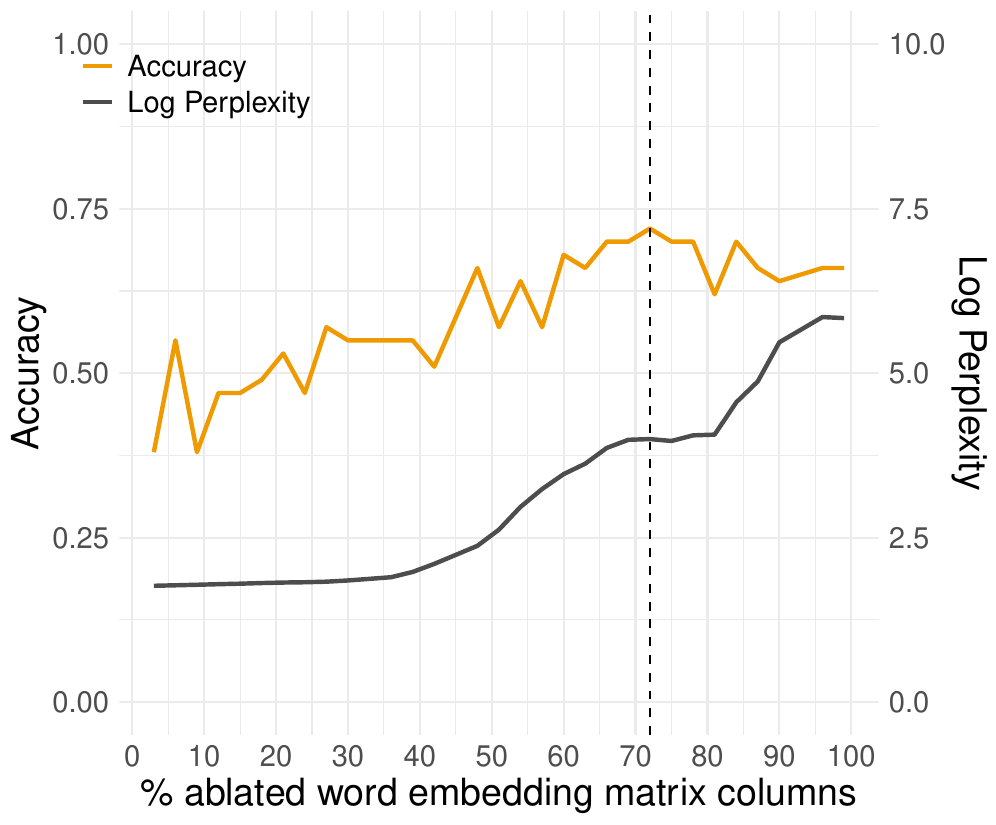}  
  \caption{GPT-2 small}
  \label{fig:gpt2-wte}
\end{subfigure}
\begin{subfigure}{.5\textwidth}
  \centering
  \includegraphics[width=.8\columnwidth]{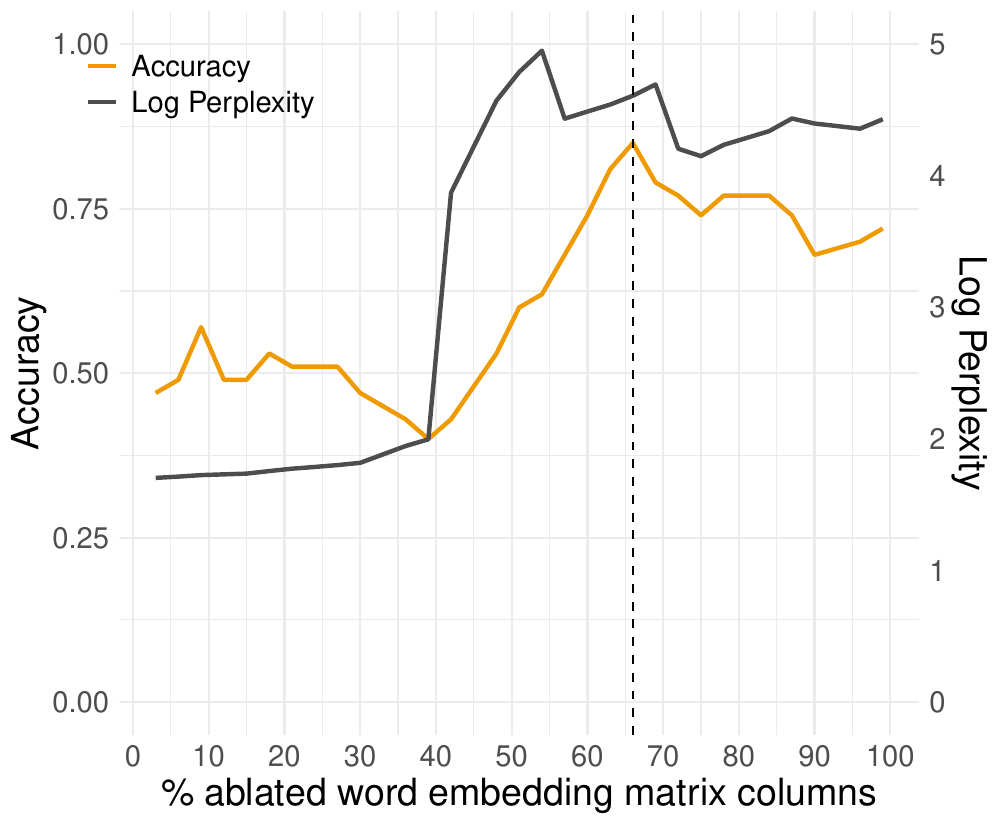}
  \caption{GPT-2 medium}
  \label{fig:gpt2-medium-wte}
\end{subfigure}
\newline
\begin{subfigure}{.5\textwidth}
  \centering
  \includegraphics[width=.8\columnwidth]{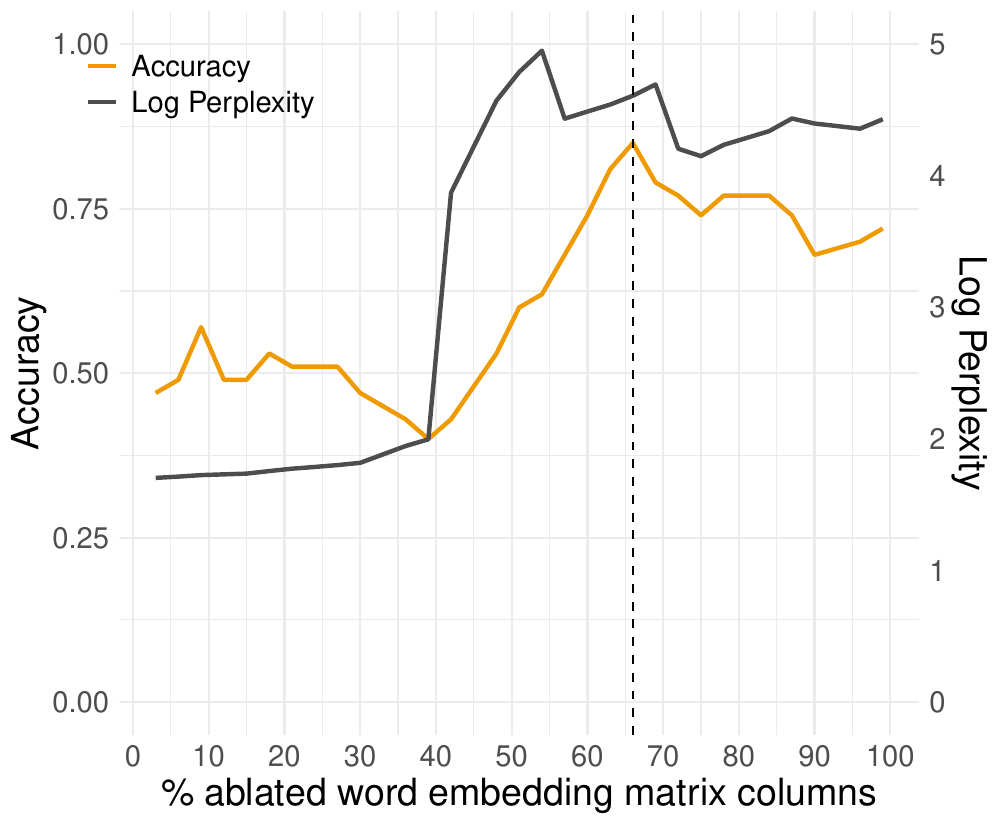}
  \caption{GPT-2 large}
  \label{fig:gpt2-large-wte}
\end{subfigure}
\begin{subfigure}{.5\textwidth}
  \centering
  \includegraphics[width=.8\columnwidth]{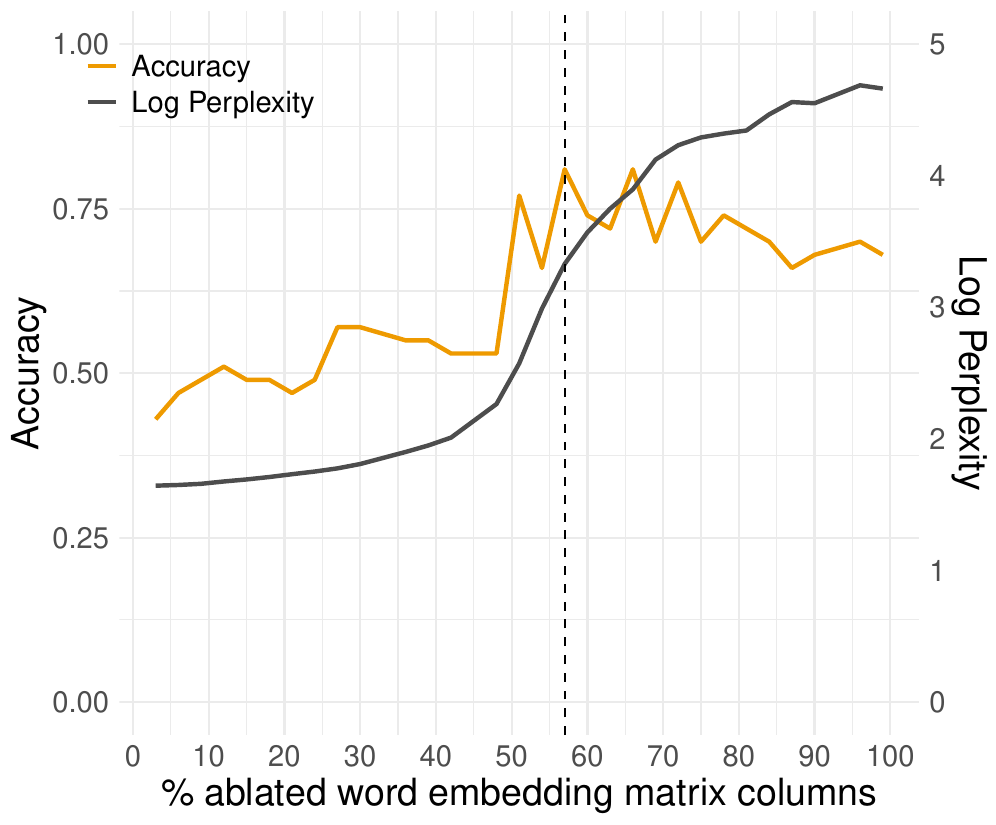}  
  \caption{GPT-2 XL}
  \label{fig:gpt2-xl-wte}
\end{subfigure}
\caption{Comparison of GPT-2 models with masked columns of word embedding matrix on classification performance and cognitive reserve manifestation. The left y-axis denotes classification performance using both masked and unmasked GPT-2 models on the ADReSS test set. The right y-axis indicates log PPL estimated from transcripts of WLS healthy individuals. The x-axis represents the percentage of attention heads getting masked. The vertical dashed line indicates the best-performing masking pattern, achieving the highest ACC.}
\label{fig:mask-wte}
\end{figure*}

\begin{table*}
\small
\centering
\begin{filecontents*}{gpt2-head-rank.txt}
121.000000 5.000000 7.000000 65.000000 19.000000 46.000000 58.000000 56.000000 9.000000 1.000000 0.000000 60.000000
16.000000 34.000000 94.000000 71.000000 13.000000 90.000000 42.000000 4.000000 113.000000 86.000000 47.000000 14.000000
74.000000 106.000000 107.000000 32.000000 89.000000 18.000000 17.000000 87.000000 75.000000 11.000000 8.000000 128.000000
48.000000 15.000000 82.000000 78.000000 93.000000 68.000000 129.000000 79.000000 77.000000 72.000000 54.000000 23.000000
40.000000 67.000000 22.000000 115.000000 36.000000 131.000000 100.000000 83.000000 140.000000 61.000000 141.000000 135.000000
41.000000 29.000000 134.000000 137.000000 119.000000 12.000000 92.000000 21.000000 31.000000 3.000000 69.000000 28.000000
76.000000 95.000000 101.000000 125.000000 91.000000 130.000000 37.000000 99.000000 80.000000 98.000000 10.000000 122.000000
117.000000 45.000000 124.000000 81.000000 116.000000 49.000000 25.000000 26.000000 62.000000 97.000000 143.000000 136.000000
64.000000 123.000000 30.000000 43.000000 88.000000 38.000000 27.000000 55.000000 73.000000 114.000000 118.000000 142.000000
111.000000 53.000000 102.000000 70.000000 50.000000 57.000000 105.000000 84.000000 120.000000 138.000000 139.000000 20.000000
132.000000 110.000000 66.000000 103.000000 44.000000 52.000000 126.000000 108.000000 109.000000 59.000000 96.000000 85.000000
6.000000 24.000000 127.000000 39.000000 33.000000 133.000000 104.000000 51.000000 2.000000 112.000000 63.000000 35.000000
\end{filecontents*}
\pgfplotstabletypeset[dec sep align,
   fixed, 
   every head row/.style={before row=\hline,after row=\hline},
   col sep=space]{gpt2-head-rank.txt}
\caption{The rank of importance for each attention head in the GPT-2 small model. The rows represent the layer of attention blocks in the model whereas the columns represent attention heads per layer.}
\label{tab:gpt2-rank}
\end{table*}

\begin{table*}
\small
\centering
\begin{filecontents*}{gpt2-medium-head-rank.txt}
139.000000 115.000000 36.000000 376.000000 354.000000 132.000000 355.000000 357.000000 109.000000 70.000000 148.000000 84.000000 217.000000 156.000000 309.000000 363.000000
77.000000 5.000000 28.000000 126.000000 41.000000 95.000000 266.000000 57.000000 286.000000 316.000000 61.000000 258.000000 59.000000 13.000000 138.000000 202.000000
268.000000 131.000000 21.000000 209.000000 367.000000 46.000000 86.000000 228.000000 45.000000 29.000000 22.000000 63.000000 48.000000 76.000000 155.000000 18.000000
170.000000 222.000000 346.000000 235.000000 38.000000 12.000000 257.000000 301.000000 172.000000 112.000000 193.000000 121.000000 188.000000 26.000000 238.000000 103.000000
294.000000 47.000000 198.000000 102.000000 315.000000 64.000000 291.000000 40.000000 246.000000 375.000000 66.000000 151.000000 292.000000 293.000000 264.000000 165.000000
199.000000 272.000000 250.000000 364.000000 285.000000 226.000000 192.000000 343.000000 17.000000 114.000000 119.000000 260.000000 237.000000 166.000000 62.000000 68.000000
50.000000 16.000000 360.000000 185.000000 20.000000 368.000000 359.000000 130.000000 350.000000 72.000000 8.000000 289.000000 267.000000 251.000000 310.000000 7.000000
279.000000 88.000000 216.000000 79.000000 141.000000 256.000000 85.000000 83.000000 101.000000 82.000000 204.000000 232.000000 243.000000 142.000000 107.000000 55.000000
53.000000 284.000000 195.000000 129.000000 253.000000 133.000000 97.000000 154.000000 137.000000 262.000000 281.000000 60.000000 203.000000 177.000000 31.000000 248.000000
58.000000 261.000000 366.000000 25.000000 135.000000 227.000000 320.000000 333.000000 197.000000 94.000000 242.000000 125.000000 65.000000 15.000000 273.000000 117.000000
255.000000 271.000000 160.000000 54.000000 49.000000 269.000000 303.000000 140.000000 176.000000 296.000000 167.000000 311.000000 110.000000 92.000000 290.000000 239.000000
143.000000 241.000000 158.000000 325.000000 74.000000 299.000000 56.000000 342.000000 287.000000 225.000000 214.000000 6.000000 372.000000 98.000000 150.000000 100.000000
183.000000 182.000000 52.000000 370.000000 19.000000 11.000000 186.000000 113.000000 240.000000 353.000000 184.000000 34.000000 312.000000 297.000000 179.000000 259.000000
32.000000 210.000000 358.000000 212.000000 331.000000 67.000000 371.000000 230.000000 330.000000 116.000000 304.000000 263.000000 159.000000 278.000000 163.000000 149.000000
87.000000 324.000000 208.000000 334.000000 356.000000 220.000000 319.000000 162.000000 136.000000 236.000000 69.000000 108.000000 327.000000 190.000000 337.000000 27.000000
383.000000 14.000000 4.000000 106.000000 300.000000 275.000000 96.000000 71.000000 207.000000 75.000000 352.000000 10.000000 221.000000 178.000000 307.000000 180.000000
339.000000 174.000000 336.000000 340.000000 44.000000 191.000000 99.000000 317.000000 39.000000 244.000000 361.000000 249.000000 274.000000 73.000000 206.000000 377.000000
362.000000 341.000000 201.000000 345.000000 231.000000 219.000000 35.000000 111.000000 205.000000 80.000000 93.000000 23.000000 347.000000 168.000000 229.000000 30.000000
378.000000 146.000000 171.000000 145.000000 3.000000 181.000000 321.000000 196.000000 124.000000 164.000000 152.000000 120.000000 224.000000 276.000000 382.000000 24.000000
33.000000 302.000000 128.000000 78.000000 252.000000 298.000000 189.000000 144.000000 344.000000 42.000000 105.000000 349.000000 329.000000 288.000000 104.000000 283.000000
369.000000 37.000000 90.000000 247.000000 295.000000 305.000000 81.000000 314.000000 318.000000 215.000000 173.000000 123.000000 313.000000 254.000000 365.000000 306.000000
277.000000 43.000000 381.000000 118.000000 373.000000 280.000000 233.000000 380.000000 234.000000 270.000000 374.000000 211.000000 335.000000 282.000000 323.000000 332.000000
157.000000 322.000000 2.000000 51.000000 326.000000 161.000000 147.000000 200.000000 218.000000 338.000000 348.000000 213.000000 379.000000 89.000000 153.000000 122.000000
187.000000 194.000000 265.000000 175.000000 134.000000 1.000000 0.000000 351.000000 169.000000 127.000000 328.000000 91.000000 308.000000 223.000000 245.000000 9.000000
\end{filecontents*}
\pgfplotstabletypeset[dec sep align,
   fixed, 
   every head row/.style={before row=\hline,after row=\hline},
   col sep=space]{gpt2-medium-head-rank.txt}
\caption{The rank of importance for each attention head in the GPT-2 medium model. The rows represent the layer of attention blocks in the model whereas the columns represent attention heads per layer.}
\label{tab:gpt2-medium-rank}
\end{table*}

\begin{table*}
\centering
\small
\resizebox{\textwidth}{!}{
\begin{filecontents*}{gpt2-large-head-rank.txt}
39.000000 79.000000 318.000000 53.000000 587.000000 121.000000 273.000000 390.000000 84.000000 125.000000 433.000000 510.000000 653.000000 548.000000 379.000000 255.000000 193.000000 132.000000 555.000000 622.000000
567.000000 285.000000 553.000000 403.000000 481.000000 670.000000 375.000000 398.000000 399.000000 354.000000 298.000000 424.000000 502.000000 624.000000 665.000000 544.000000 709.000000 473.000000 620.000000 268.000000
219.000000 181.000000 585.000000 683.000000 621.000000 463.000000 326.000000 506.000000 434.000000 615.000000 339.000000 651.000000 100.000000 312.000000 497.000000 332.000000 412.000000 428.000000 14.000000 634.000000
12.000000 22.000000 633.000000 391.000000 436.000000 658.000000 94.000000 563.000000 614.000000 386.000000 355.000000 104.000000 174.000000 40.000000 90.000000 172.000000 78.000000 407.000000 328.000000 673.000000
149.000000 302.000000 105.000000 322.000000 474.000000 264.000000 469.000000 184.000000 35.000000 319.000000 177.000000 185.000000 295.000000 203.000000 296.000000 438.000000 666.000000 489.000000 543.000000 87.000000
606.000000 323.000000 209.000000 195.000000 579.000000 204.000000 51.000000 584.000000 267.000000 569.000000 590.000000 662.000000 395.000000 55.000000 93.000000 290.000000 552.000000 159.000000 508.000000 186.000000
163.000000 425.000000 225.000000 25.000000 448.000000 107.000000 531.000000 325.000000 58.000000 27.000000 503.000000 612.000000 92.000000 559.000000 287.000000 118.000000 352.000000 275.000000 251.000000 546.000000
681.000000 592.000000 523.000000 77.000000 494.000000 56.000000 549.000000 342.000000 368.000000 611.000000 528.000000 320.000000 558.000000 47.000000 308.000000 575.000000 248.000000 383.000000 671.000000 314.000000
134.000000 250.000000 340.000000 272.000000 712.000000 396.000000 171.000000 137.000000 610.000000 430.000000 238.000000 269.000000 26.000000 604.000000 211.000000 630.000000 261.000000 133.000000 532.000000 111.000000
34.000000 240.000000 388.000000 408.000000 337.000000 44.000000 568.000000 247.000000 359.000000 411.000000 672.000000 215.000000 258.000000 198.000000 410.000000 547.000000 472.000000 525.000000 581.000000 702.000000
617.000000 500.000000 566.000000 468.000000 365.000000 311.000000 692.000000 533.000000 422.000000 699.000000 660.000000 645.000000 265.000000 131.000000 643.000000 526.000000 657.000000 36.000000 189.000000 632.000000
116.000000 146.000000 81.000000 413.000000 625.000000 284.000000 161.000000 62.000000 460.000000 588.000000 627.000000 545.000000 330.000000 88.000000 293.000000 527.000000 650.000000 71.000000 303.000000 245.000000
372.000000 299.000000 357.000000 524.000000 640.000000 685.000000 11.000000 31.000000 674.000000 346.000000 565.000000 599.000000 194.000000 564.000000 439.000000 145.000000 196.000000 97.000000 260.000000 167.000000
688.000000 175.000000 224.000000 263.000000 41.000000 239.000000 114.000000 294.000000 331.000000 609.000000 202.000000 249.000000 373.000000 120.000000 550.000000 119.000000 454.000000 246.000000 98.000000 220.000000
173.000000 310.000000 179.000000 381.000000 377.000000 103.000000 479.000000 135.000000 516.000000 475.000000 148.000000 205.000000 401.000000 443.000000 169.000000 560.000000 218.000000 556.000000 654.000000 600.000000
637.000000 432.000000 301.000000 394.000000 65.000000 156.000000 123.000000 329.000000 66.000000 70.000000 214.000000 574.000000 2.000000 10.000000 307.000000 153.000000 46.000000 112.000000 613.000000 237.000000
166.000000 542.000000 50.000000 155.000000 217.000000 117.000000 435.000000 305.000000 417.000000 43.000000 138.000000 102.000000 367.000000 327.000000 253.000000 570.000000 414.000000 343.000000 143.000000 207.000000
243.000000 647.000000 7.000000 191.000000 698.000000 109.000000 141.000000 423.000000 598.000000 126.000000 182.000000 0.000000 421.000000 17.000000 24.000000 52.000000 110.000000 80.000000 59.000000 91.000000
234.000000 488.000000 351.000000 256.000000 649.000000 113.000000 551.000000 668.000000 164.000000 400.000000 501.000000 128.000000 83.000000 282.000000 210.000000 317.000000 45.000000 511.000000 222.000000 68.000000
513.000000 274.000000 199.000000 594.000000 519.000000 348.000000 140.000000 168.000000 151.000000 157.000000 664.000000 188.000000 244.000000 449.000000 144.000000 347.000000 515.000000 324.000000 522.000000 619.000000
23.000000 122.000000 82.000000 656.000000 447.000000 358.000000 642.000000 1.000000 16.000000 42.000000 589.000000 646.000000 233.000000 466.000000 304.000000 361.000000 37.000000 32.000000 96.000000 30.000000
28.000000 216.000000 397.000000 364.000000 562.000000 29.000000 101.000000 356.000000 471.000000 695.000000 270.000000 276.000000 162.000000 283.000000 170.000000 306.000000 277.000000 5.000000 573.000000 486.000000
300.000000 493.000000 74.000000 517.000000 678.000000 402.000000 130.000000 576.000000 165.000000 459.000000 418.000000 4.000000 426.000000 442.000000 583.000000 190.000000 6.000000 291.000000 208.000000 38.000000
221.000000 69.000000 370.000000 85.000000 60.000000 178.000000 641.000000 491.000000 577.000000 682.000000 154.000000 106.000000 703.000000 392.000000 362.000000 54.000000 603.000000 99.000000 315.000000 371.000000
13.000000 349.000000 415.000000 313.000000 124.000000 427.000000 281.000000 338.000000 580.000000 420.000000 416.000000 380.000000 499.000000 644.000000 20.000000 384.000000 530.000000 183.000000 192.000000 72.000000
297.000000 286.000000 440.000000 648.000000 75.000000 201.000000 57.000000 561.000000 136.000000 152.000000 33.000000 242.000000 95.000000 108.000000 369.000000 697.000000 716.000000 229.000000 266.000000 534.000000
206.000000 477.000000 687.000000 538.000000 707.000000 127.000000 706.000000 280.000000 64.000000 409.000000 669.000000 19.000000 496.000000 200.000000 376.000000 490.000000 150.000000 498.000000 514.000000 616.000000
482.000000 158.000000 487.000000 444.000000 9.000000 160.000000 197.000000 231.000000 406.000000 76.000000 21.000000 187.000000 180.000000 437.000000 693.000000 257.000000 139.000000 467.000000 321.000000 504.000000
223.000000 462.000000 485.000000 419.000000 572.000000 591.000000 142.000000 288.000000 363.000000 350.000000 719.000000 536.000000 704.000000 718.000000 677.000000 334.000000 675.000000 680.000000 445.000000 623.000000
686.000000 405.000000 701.000000 539.000000 8.000000 596.000000 278.000000 582.000000 230.000000 456.000000 341.000000 529.000000 710.000000 241.000000 344.000000 86.000000 458.000000 652.000000 635.000000 541.000000
63.000000 89.000000 366.000000 382.000000 476.000000 232.000000 638.000000 271.000000 453.000000 235.000000 227.000000 404.000000 262.000000 73.000000 67.000000 717.000000 374.000000 228.000000 512.000000 212.000000
605.000000 601.000000 15.000000 484.000000 389.000000 3.000000 176.000000 483.000000 509.000000 667.000000 495.000000 345.000000 393.000000 289.000000 309.000000 259.000000 115.000000 387.000000 446.000000 129.000000
48.000000 465.000000 557.000000 385.000000 18.000000 535.000000 628.000000 705.000000 554.000000 602.000000 478.000000 492.000000 316.000000 360.000000 236.000000 521.000000 254.000000 684.000000 252.000000 586.000000
689.000000 571.000000 464.000000 607.000000 691.000000 292.000000 450.000000 696.000000 49.000000 520.000000 333.000000 626.000000 505.000000 713.000000 700.000000 631.000000 452.000000 593.000000 694.000000 335.000000
636.000000 578.000000 147.000000 61.000000 714.000000 279.000000 455.000000 708.000000 378.000000 655.000000 595.000000 480.000000 461.000000 679.000000 518.000000 597.000000 451.000000 659.000000 661.000000 507.000000
608.000000 540.000000 226.000000 639.000000 431.000000 336.000000 676.000000 213.000000 470.000000 715.000000 618.000000 441.000000 711.000000 690.000000 353.000000 457.000000 629.000000 537.000000 429.000000 663.000000
\end{filecontents*}
\pgfplotstabletypeset[dec sep align,
   fixed, 
   every head row/.style={before row=\hline,after row=\hline},
   col sep=space]{gpt2-large-head-rank.txt}}
\caption{The rank of importance for each attention head in the GPT-2 large model. The rows represent the layer of attention blocks in the model whereas the columns represent attention heads per layer.}
\label{tab:gpt2-large-rank}
\end{table*}

\begin{sidewaystable*}
\centering
\resizebox{\textwidth}{!}{
\begin{filecontents*}{gpt2-xl-head-rank.txt}
131.000000 329.000000 685.000000 483.000000 1163.000000 980.000000 687.000000 210.000000 590.000000 35.000000 636.000000 1138.000000 84.000000 203.000000 800.000000 611.000000 993.000000 710.000000 1135.000000 822.000000 158.000000 828.000000 122.000000 869.000000 1052.000000
1134.000000 580.000000 98.000000 849.000000 915.000000 900.000000 468.000000 564.000000 1061.000000 141.000000 367.000000 529.000000 725.000000 489.000000 961.000000 352.000000 823.000000 260.000000 1027.000000 841.000000 522.000000 547.000000 209.000000 521.000000 226.000000
1086.000000 699.000000 850.000000 735.000000 268.000000 840.000000 647.000000 1019.000000 985.000000 1068.000000 757.000000 138.000000 1028.000000 690.000000 1033.000000 762.000000 708.000000 499.000000 811.000000 635.000000 193.000000 256.000000 441.000000 720.000000 820.000000
294.000000 414.000000 753.000000 261.000000 743.000000 30.000000 244.000000 1032.000000 598.000000 388.000000 691.000000 1063.000000 62.000000 574.000000 628.000000 325.000000 1037.000000 548.000000 785.000000 538.000000 599.000000 763.000000 803.000000 380.000000 798.000000
867.000000 799.000000 524.000000 304.000000 223.000000 858.000000 85.000000 814.000000 937.000000 911.000000 646.000000 838.000000 614.000000 106.000000 61.000000 66.000000 133.000000 172.000000 836.000000 173.000000 427.000000 262.000000 144.000000 1085.000000 749.000000
931.000000 342.000000 276.000000 1056.000000 1048.000000 686.000000 531.000000 101.000000 904.000000 116.000000 1073.000000 923.000000 295.000000 550.000000 880.000000 376.000000 473.000000 1148.000000 429.000000 501.000000 410.000000 422.000000 313.000000 587.000000 895.000000
802.000000 862.000000 257.000000 853.000000 1067.000000 1174.000000 280.000000 661.000000 563.000000 208.000000 1003.000000 946.000000 950.000000 26.000000 702.000000 629.000000 813.000000 440.000000 252.000000 361.000000 975.000000 493.000000 660.000000 1112.000000 790.000000
568.000000 818.000000 609.000000 321.000000 772.000000 1065.000000 196.000000 1087.000000 847.000000 1092.000000 1004.000000 945.000000 165.000000 758.000000 246.000000 949.000000 553.000000 1108.000000 387.000000 963.000000 582.000000 285.000000 572.000000 876.000000 514.000000
341.000000 891.000000 123.000000 509.000000 195.000000 269.000000 697.000000 404.000000 534.000000 666.000000 825.000000 731.000000 1115.000000 1079.000000 199.000000 463.000000 701.000000 674.000000 546.000000 642.000000 667.000000 936.000000 736.000000 1069.000000 781.000000
567.000000 1013.000000 274.000000 600.000000 353.000000 864.000000 496.000000 637.000000 729.000000 922.000000 604.000000 952.000000 976.000000 1066.000000 474.000000 754.000000 791.000000 874.000000 999.000000 318.000000 243.000000 1055.000000 909.000000 832.000000 328.000000
556.000000 623.000000 1006.000000 767.000000 1076.000000 594.000000 998.000000 833.000000 616.000000 643.000000 222.000000 1075.000000 776.000000 227.000000 1043.000000 535.000000 877.000000 902.000000 349.000000 752.000000 533.000000 494.000000 436.000000 815.000000 301.000000
1024.000000 296.000000 288.000000 603.000000 898.000000 613.000000 872.000000 1090.000000 334.000000 482.000000 1040.000000 419.000000 537.000000 1094.000000 464.000000 1042.000000 982.000000 1050.000000 719.000000 507.000000 236.000000 644.000000 459.000000 395.000000 504.000000
1143.000000 467.000000 391.000000 694.000000 1136.000000 848.000000 924.000000 656.000000 372.000000 442.000000 300.000000 444.000000 1016.000000 715.000000 184.000000 491.000000 997.000000 392.000000 751.000000 1147.000000 801.000000 174.000000 511.000000 704.000000 330.000000
323.000000 851.000000 189.000000 204.000000 322.000000 968.000000 78.000000 525.000000 714.000000 27.000000 544.000000 416.000000 439.000000 1054.000000 552.000000 727.000000 1078.000000 559.000000 804.000000 761.000000 506.000000 1118.000000 215.000000 390.000000 358.000000
816.000000 1038.000000 118.000000 162.000000 38.000000 381.000000 485.000000 201.000000 309.000000 368.000000 933.000000 95.000000 284.000000 183.000000 760.000000 396.000000 889.000000 755.000000 82.000000 649.000000 111.000000 282.000000 446.000000 455.000000 221.000000
1164.000000 335.000000 806.000000 454.000000 912.000000 247.000000 943.000000 59.000000 449.000000 756.000000 526.000000 48.000000 402.000000 1121.000000 919.000000 212.000000 412.000000 837.000000 684.000000 277.000000 497.000000 156.000000 819.000000 831.000000 437.000000
103.000000 782.000000 86.000000 3.000000 586.000000 665.000000 827.000000 595.000000 765.000000 695.000000 431.000000 766.000000 187.000000 100.000000 348.000000 167.000000 344.000000 709.000000 645.000000 1074.000000 251.000000 71.000000 640.000000 479.000000 558.000000
224.000000 592.000000 216.000000 1002.000000 161.000000 658.000000 308.000000 887.000000 555.000000 420.000000 317.000000 258.000000 995.000000 528.000000 339.000000 46.000000 153.000000 302.000000 1011.000000 265.000000 255.000000 272.000000 241.000000 99.000000 75.000000
530.000000 676.000000 428.000000 516.000000 808.000000 305.000000 712.000000 64.000000 705.000000 155.000000 225.000000 393.000000 76.000000 596.000000 287.000000 397.000000 69.000000 92.000000 137.000000 220.000000 634.000000 266.000000 360.000000 430.000000 248.000000
927.000000 659.000000 333.000000 458.000000 857.000000 0.000000 235.000000 242.000000 978.000000 424.000000 117.000000 389.000000 185.000000 120.000000 14.000000 351.000000 44.000000 878.000000 996.000000 152.000000 830.000000 487.000000 654.000000 905.000000 128.000000
942.000000 168.000000 477.000000 232.000000 698.000000 855.000000 602.000000 178.000000 495.000000 386.000000 327.000000 177.000000 51.000000 2.000000 1045.000000 331.000000 88.000000 519.000000 403.000000 413.000000 81.000000 132.000000 608.000000 369.000000 470.000000
1035.000000 575.000000 110.000000 213.000000 157.000000 39.000000 707.000000 1039.000000 139.000000 1126.000000 466.000000 411.000000 211.000000 121.000000 638.000000 316.000000 45.000000 159.000000 1072.000000 515.000000 1071.000000 633.000000 175.000000 539.000000 49.000000
956.000000 401.000000 96.000000 597.000000 721.000000 965.000000 663.000000 160.000000 217.000000 36.000000 768.000000 143.000000 472.000000 194.000000 320.000000 56.000000 170.000000 297.000000 25.000000 113.000000 696.000000 984.000000 566.000000 23.000000 681.000000
283.000000 462.000000 53.000000 40.000000 517.000000 127.000000 626.000000 125.000000 12.000000 291.000000 83.000000 1.000000 1020.000000 55.000000 33.000000 443.000000 310.000000 286.000000 500.000000 958.000000 150.000000 65.000000 197.000000 108.000000 839.000000
778.000000 682.000000 10.000000 332.000000 693.000000 114.000000 180.000000 679.000000 346.000000 383.000000 169.000000 90.000000 289.000000 237.000000 706.000000 182.000000 503.000000 1009.000000 347.000000 480.000000 861.000000 319.000000 1137.000000 377.000000 119.000000
783.000000 409.000000 171.000000 202.000000 452.000000 16.000000 130.000000 856.000000 164.000000 303.000000 565.000000 866.000000 253.000000 4.000000 475.000000 1144.000000 250.000000 451.000000 773.000000 711.000000 19.000000 1053.000000 1021.000000 746.000000 1093.000000
6.000000 275.000000 967.000000 1156.000000 908.000000 15.000000 29.000000 188.000000 780.000000 415.000000 1060.000000 206.000000 24.000000 846.000000 47.000000 668.000000 200.000000 536.000000 281.000000 577.000000 860.000000 615.000000 1064.000000 883.000000 438.000000
104.000000 703.000000 270.000000 399.000000 94.000000 179.000000 234.000000 1057.000000 578.000000 759.000000 77.000000 445.000000 627.000000 371.000000 884.000000 733.000000 50.000000 518.000000 264.000000 1104.000000 1034.000000 523.000000 826.000000 1123.000000 80.000000
9.000000 621.000000 142.000000 146.000000 354.000000 312.000000 363.000000 238.000000 259.000000 1023.000000 935.000000 450.000000 1184.000000 239.000000 407.000000 1102.000000 73.000000 357.000000 741.000000 425.000000 207.000000 307.000000 769.000000 89.000000 421.000000
292.000000 576.000000 22.000000 109.000000 688.000000 405.000000 540.000000 214.000000 988.000000 983.000000 770.000000 498.000000 293.000000 896.000000 810.000000 794.000000 271.000000 63.000000 1081.000000 28.000000 398.000000 744.000000 502.000000 70.000000 914.000000
240.000000 879.000000 508.000000 1036.000000 151.000000 1103.000000 364.000000 58.000000 433.000000 492.000000 112.000000 434.000000 8.000000 873.000000 350.000000 1195.000000 871.000000 273.000000 1180.000000 417.000000 52.000000 1141.000000 617.000000 1175.000000 620.000000
1167.000000 954.000000 1110.000000 510.000000 135.000000 1172.000000 105.000000 724.000000 870.000000 31.000000 892.000000 298.000000 824.000000 692.000000 340.000000 657.000000 554.000000 994.000000 486.000000 315.000000 562.000000 717.000000 115.000000 974.000000 448.000000
1015.000000 934.000000 91.000000 549.000000 1159.000000 581.000000 453.000000 356.000000 835.000000 418.000000 732.000000 513.000000 888.000000 102.000000 5.000000 488.000000 490.000000 906.000000 859.000000 1140.000000 97.000000 447.000000 570.000000 43.000000 821.000000
1031.000000 481.000000 1130.000000 456.000000 1193.000000 149.000000 32.000000 683.000000 734.000000 166.000000 1162.000000 205.000000 573.000000 370.000000 355.000000 233.000000 951.000000 1139.000000 817.000000 662.000000 147.000000 561.000000 700.000000 219.000000 972.000000
903.000000 1129.000000 136.000000 986.000000 560.000000 72.000000 1154.000000 716.000000 957.000000 41.000000 672.000000 625.000000 190.000000 973.000000 145.000000 1089.000000 894.000000 605.000000 1181.000000 916.000000 1082.000000 218.000000 469.000000 484.000000 42.000000
1128.000000 932.000000 579.000000 362.000000 385.000000 249.000000 793.000000 1083.000000 962.000000 723.000000 788.000000 435.000000 632.000000 1150.000000 601.000000 1047.000000 1091.000000 229.000000 34.000000 738.000000 326.000000 191.000000 730.000000 532.000000 588.000000
465.000000 1005.000000 1149.000000 722.000000 882.000000 829.000000 1099.000000 1151.000000 11.000000 359.000000 20.000000 792.000000 797.000000 1131.000000 737.000000 726.000000 148.000000 1098.000000 718.000000 198.000000 365.000000 87.000000 713.000000 299.000000 675.000000
1022.000000 471.000000 461.000000 163.000000 610.000000 655.000000 337.000000 1194.000000 1077.000000 991.000000 1044.000000 1179.000000 57.000000 1012.000000 426.000000 777.000000 1018.000000 545.000000 1170.000000 750.000000 1176.000000 126.000000 886.000000 901.000000 343.000000
669.000000 408.000000 140.000000 928.000000 378.000000 779.000000 254.000000 899.000000 947.000000 314.000000 382.000000 520.000000 324.000000 1111.000000 955.000000 1095.000000 622.000000 21.000000 338.000000 893.000000 278.000000 689.000000 311.000000 925.000000 107.000000
673.000000 263.000000 1029.000000 748.000000 1158.000000 60.000000 940.000000 1192.000000 742.000000 843.000000 1058.000000 345.000000 457.000000 966.000000 868.000000 854.000000 1025.000000 366.000000 245.000000 591.000000 885.000000 787.000000 929.000000 740.000000 680.000000
1008.000000 186.000000 1169.000000 154.000000 1124.000000 1062.000000 541.000000 960.000000 379.000000 181.000000 653.000000 897.000000 1197.000000 1000.000000 1046.000000 1096.000000 786.000000 1101.000000 807.000000 306.000000 845.000000 569.000000 67.000000 1119.000000 959.000000
1155.000000 54.000000 969.000000 18.000000 970.000000 863.000000 964.000000 989.000000 93.000000 739.000000 13.000000 842.000000 375.000000 423.000000 651.000000 917.000000 374.000000 1105.000000 1030.000000 907.000000 987.000000 812.000000 1173.000000 795.000000 583.000000
17.000000 589.000000 279.000000 1185.000000 938.000000 134.000000 1145.000000 944.000000 593.000000 1117.000000 747.000000 571.000000 875.000000 913.000000 267.000000 948.000000 37.000000 68.000000 630.000000 400.000000 129.000000 585.000000 124.000000 1041.000000 1166.000000
910.000000 1153.000000 1182.000000 920.000000 1160.000000 671.000000 1199.000000 618.000000 1109.000000 639.000000 74.000000 1188.000000 527.000000 881.000000 1049.000000 612.000000 771.000000 953.000000 1088.000000 1196.000000 1168.000000 1152.000000 290.000000 926.000000 764.000000
1165.000000 1100.000000 512.000000 844.000000 1059.000000 784.000000 336.000000 1107.000000 1120.000000 852.000000 992.000000 641.000000 1127.000000 670.000000 981.000000 1178.000000 1001.000000 373.000000 921.000000 230.000000 918.000000 939.000000 1171.000000 1183.000000 1187.000000
789.000000 1014.000000 1198.000000 228.000000 631.000000 551.000000 930.000000 971.000000 460.000000 432.000000 1122.000000 1070.000000 406.000000 1142.000000 774.000000 619.000000 805.000000 557.000000 728.000000 607.000000 678.000000 1113.000000 796.000000 1191.000000 1017.000000
1026.000000 584.000000 1007.000000 648.000000 941.000000 476.000000 1133.000000 650.000000 1186.000000 624.000000 1190.000000 1132.000000 1114.000000 775.000000 384.000000 1051.000000 1189.000000 1125.000000 1106.000000 1010.000000 1097.000000 1161.000000 1080.000000 990.000000 890.000000
79.000000 1157.000000 652.000000 231.000000 1177.000000 176.000000 542.000000 192.000000 1116.000000 1146.000000 505.000000 478.000000 745.000000 977.000000 394.000000 677.000000 664.000000 865.000000 979.000000 543.000000 606.000000 1084.000000 7.000000 809.000000 834.000000
\end{filecontents*}
\pgfplotstabletypeset[dec sep align,
   fixed, 
   every head row/.style={before row=\hline,after row=\hline},
   col sep=space]{gpt2-xl-head-rank.txt}}
\caption{The rank of importance for each attention head in the GPT-2 XL model. The rows represent the layer of attention blocks in the model whereas the columns represent attention heads per layer.}
\label{tab:gpt2-xl-rank}
\end{sidewaystable*}

\end{document}